\documentclass[10pt,twocolumn,letterpaper]{article}

\usepackage[pagenumbers]{cvpr} % To force page numbers, e.g. for an arXiv version

\usepackage{multirow} 
\usepackage{makecell}
\usepackage{enumitem}

\definecolor{cvprblue}{rgb}{0.21,0.49,0.74}
\definecolor{mygreen}{RGB}{0, 128, 0}

\usepackage[pagebackref,breaklinks,colorlinks,allcolors=cvprblue]{hyperref}

\def\paperID{6802} % *** Enter the Paper ID here
\def\confName{CVPR}
\def\confYear{2025}

\title{Joint Vision-Language Social Bias Removal for CLIP}

\author{
$\text{Haoyu Zhang}$, 
$\text{Yangyang Guo}$, 
$\text{Mohan Kankanhalli}$\\
$\text{National University of Singapore}$
}

\begin{document}
\maketitle
\begin{abstract}
Vision-Language (V-L) pre-trained models such as CLIP show prominent capabilities in various downstream tasks. Despite this promise, V-L models are notoriously limited by their inherent social biases. A typical demonstration is that V-L models often produce biased predictions against specific groups of people, significantly undermining their real-world applicability. Existing approaches endeavor to mitigate the social bias problem in V-L models by removing biased attribute information from model embeddings. However, after our revisiting of these methods, we find that their bias removal is frequently accompanied by greatly compromised V-L alignment capabilities. We then reveal that this performance degradation stems from the unbalanced debiasing in image and text embeddings. To address this issue, we propose a novel V-L debiasing framework to align image and text biases followed by removing them from both modalities. By doing so, our method achieves multi-modal bias mitigation while maintaining the V-L alignment in the debiased embeddings. Additionally, we advocate a new evaluation protocol that can 1) holistically quantify the model debiasing and V-L alignment ability, and 2) evaluate the generalization of social bias removal models. We believe this work will offer new insights and guidance for future studies addressing the social bias problem in CLIP. Our code can be found at \href{https://github.com/haoyusimon/VL_Debiasing}{link}.
\end{abstract}    
\section{Introduction}
\label{sec:intro}
Recently, Vision-Language (V-L) pre-trained models such as CLIP~\cite{Radford-2021-clip} and BLIP~\cite{junnan-2022-blip} have gained dominant popularity. The pre-training using large-scale image-text pairs from the web aids the alignment between visual and linguistic semantics. The alignment has led to remarkable zero-shot performance across a diverse range of applications such as image classification~\cite{deng-2009-imagenet} and cross-modal retrieval~\cite{plummer-2015-flickr}. Despite this capability, a pressing issue of social bias significantly impedes V-L models' real-world deployment. In particular, these V-L models frequently develop associations of neutral concepts with people's sensitive attributes, leading to biased model outputs against specific social groups~\cite{wang-etal-2021-gender}. This problem, as hidden in their learned embeddings, primarily originates from the stereotypes~\cite{birhane-2021-multimodal} and spurious correlations~\cite{singla-2022-core} in the training data.

\begin{figure}[t]
\centering
\includegraphics[width=0.8\linewidth]{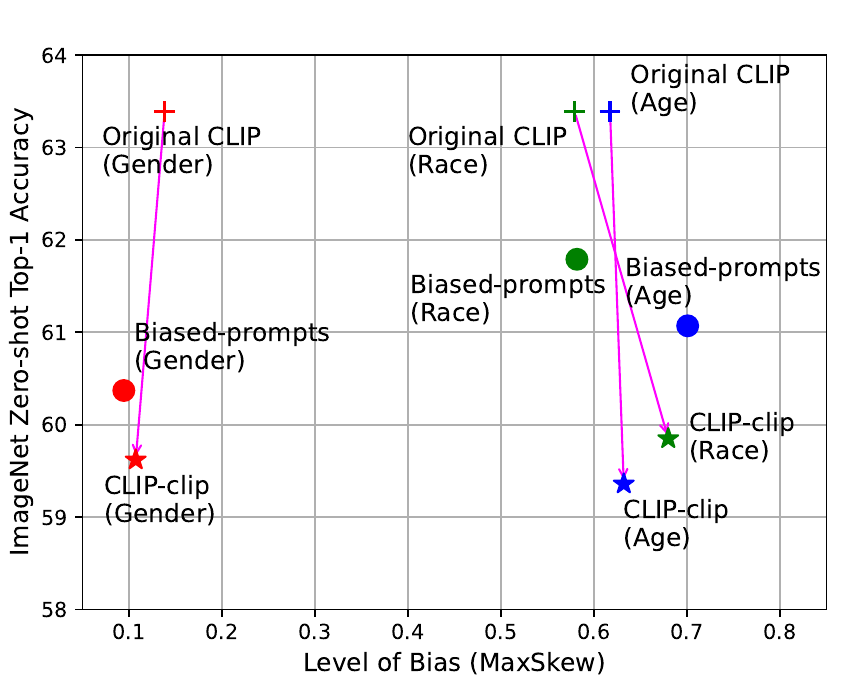} 
\vspace*{-0.3cm}
% \hspace{0.1cm}
\caption{Model results before and after removing social biases related to {\color{red}gender}, {\textcolor{mygreen}{race}}, and {\color{blue}age}, respectively.}
\vspace*{-0.7cm}
\label{fig:vl_align}
\end{figure}

Existing social debiasing\footnote{In this paper, we use the terms social debiasing and social bias removal interchangeably.} methods attempt to untie the concept-attribute associations by removing attribute information from image or/and text embeddings. They involve either learning a fair module~\cite{berg-etal-2022-prompt,seth-2023-dear, hirota-2024-saner, kong-2024-ttbias} or directly manipulating the embedding vectors (through projection~\cite{chuang-2023-debiasing, dehdashtian-2024-fairerclip, gerych-2024-bendvlm} or feature clipping based on mutual information~\cite{wang-etal-2021-gender}). However, after revisiting several representative approaches, we uncover one significant drawback: these debiasing approaches greatly compromise the V-L alignment ability in their produced embeddings - a phenomenon we term \emph{over-debiasing}. Fig.~\ref{fig:vl_align} illustrates the model results of two social bias removal methods: Biased-prompts~\cite{chuang-2023-debiasing} and CLIP-clip~\cite{wang-etal-2021-gender}. One can see that despite the alleviation of social bias level\footnote{A smaller MaxSkew indicates less social bias in models.}, the downstream model performance deteriorates significantly as indicated by the {\color{magenta}purple} arrows.

We then delve into the underlying rationale of the over-debiasing problem. Our first observation is that social biases exist in both image and text embeddings of the CLIP model (refer to Fig.~\ref{fig:vis} for more details). Nevertheless, most previous approaches debias one modality only, leaving the other one untouched~\cite{berg-etal-2022-prompt,chuang-2023-debiasing,seth-2023-dear}. CLIP-clip~\cite{wang-etal-2021-gender} attempts to debias both modalities by removing the same biased dimensions of image and text embedding vectors. However, as indicated in our second observation (Sect.~\ref{sec:3.2}), removing the exact same dimensions of text embeddings as those of vision in CLIP-clip fails to reliably debias as the image and text biases are often misaligned.

Motivated by the above observations, in this paper, we propose a novel V-L debiasing method to jointly debias both image and text modalities. Our proposed method involves a bias alignment and a counterfactual debiasing operation. Specifically, the bias alignment operation aligns the bias embeddings between image and text before debiasing. This distinguishes our method from CLIP-clip which applies the same debiasing to both modalities without alignment. Our second contribution involves triplets consisting of two oppositely biased text prompts with the same neutral information and one biased image. Under this context, we design a counterfactual debiasing objective that pulls debiased text prompts closer to remove biases while preserving the neutral information. Our method is also enabled to maintain the V-L alignment in debiased embeddings by combining debiasing with the original contrastive learning objectives. 

Additionally, we develop two new evaluation protocols for social bias removal in CLIP. We first design a new metric called Alignment and Bias Level Evaluation (ABLE), holistically evaluating the effectiveness of model debiasing and downstream V-L alignment. Thereafter, we propose to leverage out-of-domain datasets to test the generalization ability of social debiasing models, in contrast to \textit{most previous approaches using in-domain evaluation only}. When comparing with several baselines, we observe that our proposed method demonstrates significant advantages across various backbones including ViT-B/16, ViT-B/32, ViT-L/14, and ViT-H/14. Furthermore, our method allows for joint debiasing of multiple types of social biases, making it more applicable to real-world debiasing requirements.

% We summarize the contributions below:
In summary, this paper makes the following three contributions:
\begin{itemize}[leftmargin=8pt]
\setlength\itemsep{0em}
    \item We revisit existing social debiasing methods in CLIP and underline their over-debiasing problem. The innate reason for this problem is further illuminated from the perspective of misaligned V-L debiasing.
    \item We propose a novel joint V-L debiasing method to address the over-debiasing problem. Our method effectively mitigates both image and text biases despite their different distributions while maintaining V-L alignment in the debiased embeddings.
    \item We advocate two new evaluation protocols to comprehensively evaluate the bias reduction and V-L alignment in a social debiasing method.

\end{itemize}

\section{Related Work}
\label{sec:related_work}

\noindent \textbf{Social Bias Measurement.} The social bias problem (prediction inclining to specific groups of gender, age, and race) has been long a pressing concern in various realms including vision-centric, language-centric, and V-L models. In the vision domain, social biases are measured in terms of fairness metrics such as Demographic Parity~\cite{feldman-2015-dp} and Equalized Odds~\cite{hardt-2016-eo}. These metrics evaluate whether the model performance in downstream tasks, e.g., classification, varies significantly when applied to different social groups. Pertaining to language models, early work probes social biases in word-level embeddings using the World Embedding Association Test (WEAT)~\cite{caliskan-2017-weat} and projection on bias direction~\cite{bolu-2016-word_projection}. Later, the development of pre-trained language models~\cite{devlin-etal-2019-bert,radford-2019-gpt} motivates the study on the sentence-level social bias, leading to benchmarks such as Sentence Encoder Association Test (SEAT)~\cite{may-etal-2019-seat}, SteroSet~\cite{nadeem-etal-2021-stereoset} and CrowS-Pairs~\cite{nangia-etal-2020-crows}. Additionally, for V-L models like CLIP, fairness is commonly evaluated based on the proportion of retrieved images within different groups~\cite{geyik-2019-linkedin,wang-etal-2021-gender} or the diversity of the images produced by generative models conditioned on CLIP text embeddings~\cite{chuang-2023-debiasing}. Some work~\cite{Dehdashtian_2024_CVPR} also characterizes accuracy-fairness trade-offs of CLIP.

\noindent \textbf{Social Bias Removal.} To alleviate social bias problems in vision models, existing approaches typically add debiasing objectives~\cite{alvi-2019-confusion_loss,tianlu-2019-adversarial} or learn an additional fair module~\cite{mengnan-2021-fairness_module,kim-2019-multiacc}. Unlike those in the vision-centric domain, social debiasing on language models focuses more on word-level and sentence-level embeddings. For instance, Hard-Debiasing~\cite{bolu-2016-word_projection} initially identifies a bias subspace on word embeddings. The biases are projected onto this subspace and subsequently removed. Similar projection-based methods such as SentenceDebias~\cite{liang-etal-2020-sentDebias} and Iterative Nullspace Projection~\cite{ravfogel-etal-2020-inlp} along with other methods~\cite{pengyu-2021-fairfil,schick-2021-self-debias,kellie-2021-dropout_debias} are developed to remove sentence-level biases. In the V-L domain, existing CLIP social debiasing methods remove the sensitive attribute information from image and text embeddings either through direct vector manipulation or training a fair module. Vector manipulation methods involve applying a projection with closed-form solutions~\cite{chuang-2023-debiasing, dehdashtian-2024-fairerclip, gerych-2024-bendvlm} or removing features sharing the largest mutual information with sensitive attribute labels~\cite{wang-etal-2021-gender}. Differently, other approaches train a fair module with an adversarial loss~\cite{berg-etal-2022-prompt} or bias-neutralising objectives~\cite{wang-2022-fairclip, kong-2024-ttbias, hirota-2024-saner}, mostly using fairness datasets consisting of annotated images with different attributes such as FairFace \cite{kar-2021-fairface} and UTKFace \cite{zhang-2017-utkface}. Most CLIP social debiasing approaches focus on only one modality. It is worth noting that CLIP-clip~\cite{wang-etal-2021-gender} attempts multi-modal debiasing. Specifically, CLIP-clip first locates the dimensions in image embedding vectors which are mostly biased based on the mutual information between these dimensions and bias attributes. It then removes these dimensions in image embeddings and their corresponding dimensions in text embeddings. This operation assumes that dimensions containing more biases in images are also biased in the corresponding text embeddings, which, however, appears less convincing as indicated in our empirical findings.

\noindent \textbf{General Bias Removal} emphasises on addressing more general biases in models. It involves strategies to minimize spurious correlations between predictions and specific attributes in the training data to achieve group robustness. Such attributes include visual contexts in vision~\cite{polina-2023-lastlayer,liu-2021-traintwice,tsir-2023-withoutgroup} and V-L~\cite{chuang-2023-debiasing,zhang-2022-adapter, zhu-2023-finetune, Wortsman_2022_CVPR} domains or syntactic features in language tasks~\cite{mccoy-etal-2019-right,niven-kao-2019-probing,zhong-etal-2022-reducing}. Some other bias removal methods also study the long-tail problem to mitigate the bias towards dominant classes caused by class imbalance in datasets~\cite{li-2021-longtailaug,aditya-2021-logitadj,nan-etal-2021-uncovering,zhang-2022-longtail}.

\section{CLIP Social Bias Probe}
\subsection{Biases in Dual Modalities}
Our real world is unfortunately biased. Collecting interleaved image-text data from the web thus inherits its biased nature~\cite{birhane-2021-multimodal, Garcia_2023_CVPR}. As a result, pre-training on such data makes the CLIP model sensitive to certain attributes, such as gender and age~\cite{agarwal-2021-evaluatingclip, wolfe-2022-evidence}. Nevertheless, these biases appear not only in the visual world but also in our linguistic habits~\cite{wang-etal-2022-assessing}. To highlight this fact, we first study the social biases in two modalities of CLIP separately. 

\begin{figure}[t]
    \centering
        % \hspace{-0.55cm}
        \includegraphics[width=1.0\linewidth]{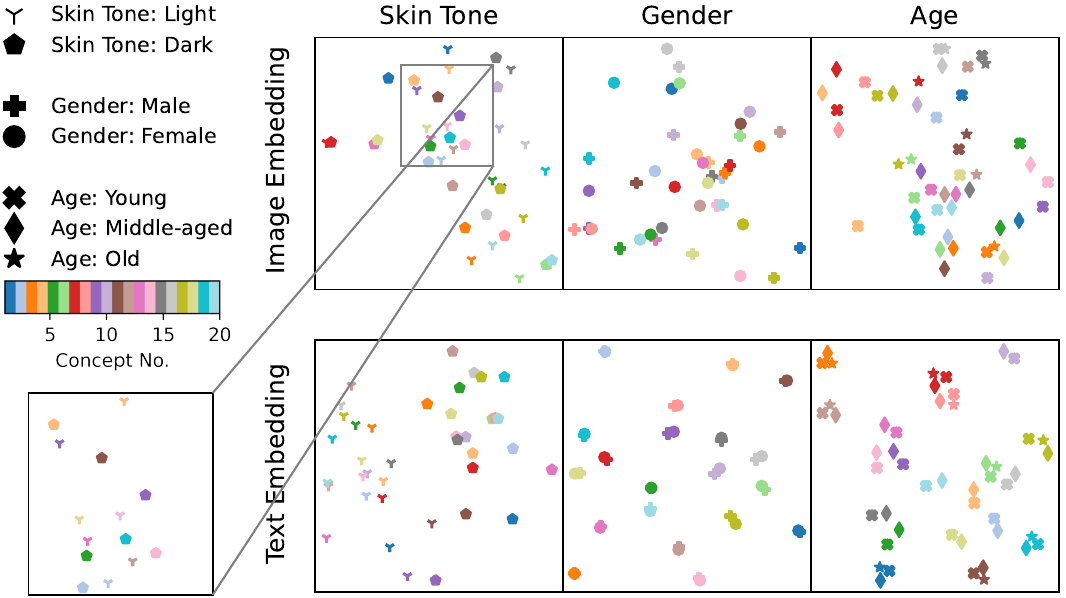}
        \vspace{-1.5em}
        \caption{Visualization of different social biases in the image (the top row) and text (the bottom row) embeddings through t-SNE. A fair model should embed different attributes (different symbols pertaining to one concept category) with respect to one concept (same color) close to each other.}
        \vspace{-1.0em}
        \label{fig:vis}
\end{figure}

We conduct experiments by the visualization of embeddings of biased text or image prompts. Specifically, we first define several biased prompt sets covering social biases like gender, age and skin tone. Each set of biased prompts consists of prompts with the same {\color{blue}neutral concept} (\eg, occupations) but different {\color{red}social attributes} (\eg, gender). For example, one of the biased text prompt sets we use is \{``A photo of a {\color{red}male} {\color{blue}dancer}'', ``A photo of a {\color{red}female} {\color{blue}dancer}''\}. For biased image prompts, we leverage the contrasting image pairs from FACET~\cite{gusta-2023-facet}, such as: \{\raisebox{-3pt}{{\includegraphics[height=15pt]{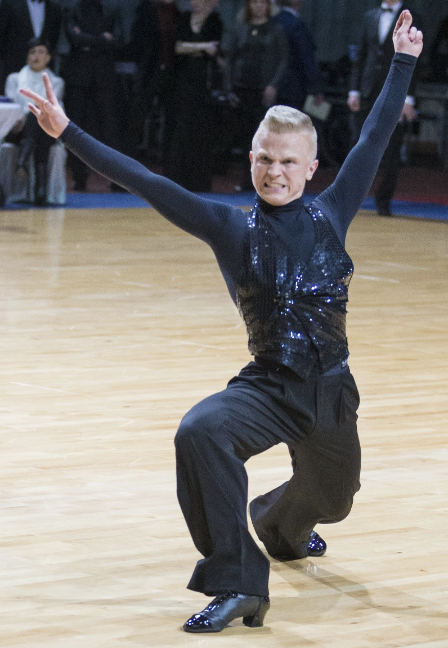}}} (a photo of a {\color{red}male} {\color{blue}dancer}), \raisebox{-3pt}{{\includegraphics[height=15pt]{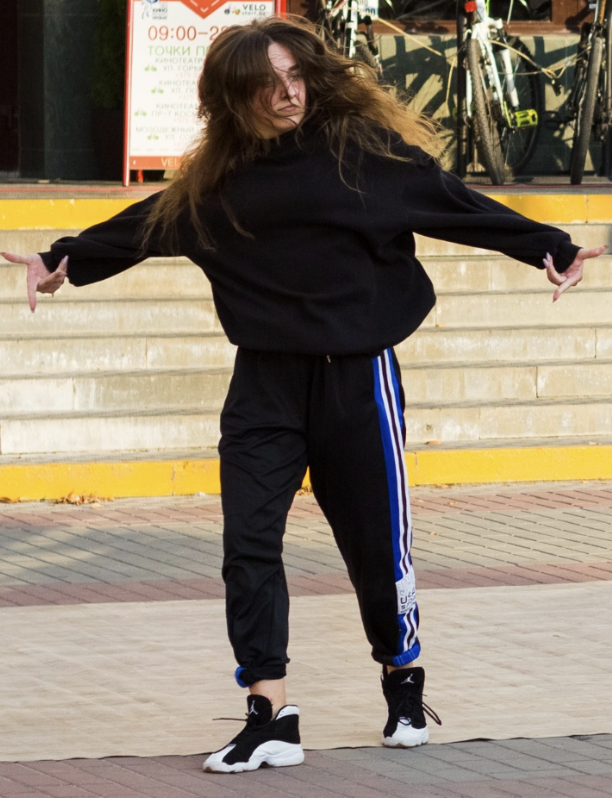}}} (a photo of a {\color{red}female} {\color{blue}dancer})\}. We employ multiple neutral concepts to obtain a diverse set of biased prompts. Thereafter, we input these image and text prompts to a pre-trained CLIP model and obtain their embeddings. By visualizing these embeddings using t-SNE in Fig.~\ref{fig:vis}, we find that social biases exist in both text and image modalities. Specifically, most image embedding pairs/triplets with the same concepts but different attributes are distributed far from each other (refer to the zoomed area of Fig.~\ref{fig:vis}). This suggests that CLIP fails to group image embeddings solely based on neutral concepts but is influenced by biases in images when encoding. On the other hand, across all three text biases, the skin tone bias is the most evident one as the text embeddings form two clusters based on different skin tones.

\noindent \textbf{Summary.} The majority of existing CLIP debiasing approaches focus on only one modality. As indicated in our experiments, there exist social biases in both text and image modalities. Under this context, debiasing one modality leads to the misalignment of image and text embeddings. The downstream task thus bears a performance degradation compared to the original CLIP models.

\subsection{Biases Are Different Across Modalities} \label{sec:3.2}
Another interesting finding from Fig.~\ref{fig:vis} is that social biases in the CLIP model are different for image and text modalities. To further elucidate this observation, we employ embedding association tests from SEAT~\cite{may-etal-2019-seat} and IEAT~\cite{steed-2021-ieat} which are widely used for evaluating social biases in text and image embeddings, respectively.

We select six common types of social bias from SEAT and IEAT. Beyond gender, age and skin tone, these tests also cover additional social biases such as weight and race. Each type of social bias is measured based on the preferential association of \textbf{two} contrasting {\color{red}sensitive attributes} with \textbf{two} contrasting {\color{blue}attribute-neutral concepts} in image or text embeddings. The {\color{red}sensitive attribute pairs} are typical bias directions such as \{``male'', ``female''\}. 
\begin{figure}
\centering
\includegraphics[width=0.9\linewidth]{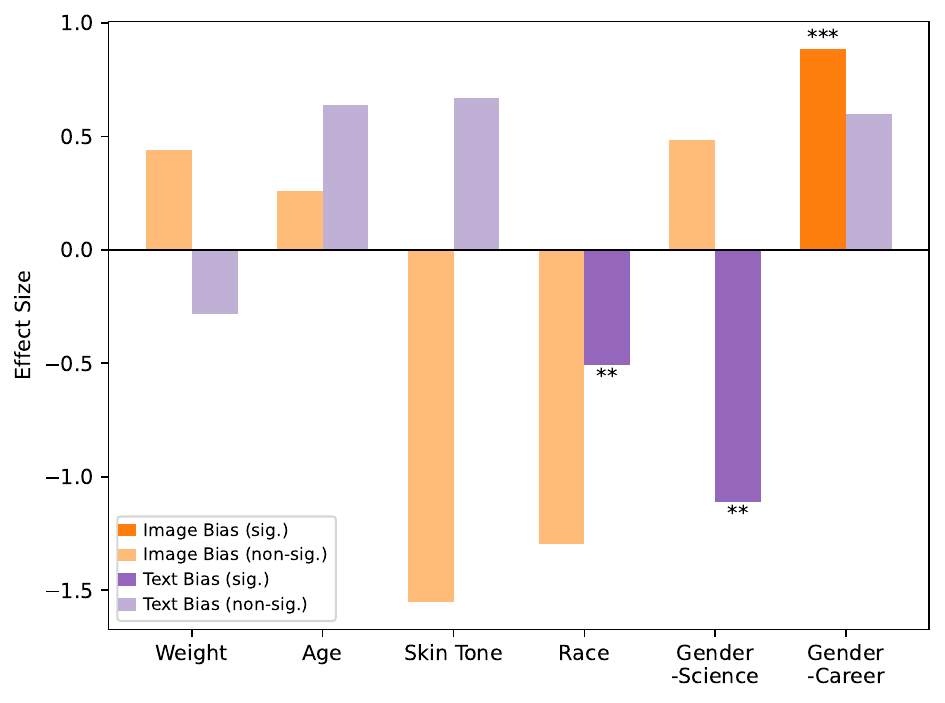}
\vspace{-1.2em}
\caption{Effective size results for text and image biases. Statistically significant (sig.) results are marked with dark blue and dark green colors. The $*$/$**$/$*$$*$$*$ implies p-values smaller than 0.1/0.05/0.01, respectively.}
\vspace{-1.2em}
\label{fig:eat}
\end{figure}The {\color{blue}neutral concept pairs} can be specific concepts such as \{``career'', ``family''\} or \{``science'', ``liberal arts''\} used in the Gender-Career and Gender-Science test or more general concepts such as \{``pleasant'', ``unpleasant''\} used in all other tests (refer to the supplementary material for more details).

The magnitudes and directions of the preferential associations (social biases) between {\color{red}attribute pairs} and {\color{blue}concept pairs} are quantified by calculating their effect sizes~\cite{caliskan-2017-weat,steed-2021-ieat}. Specifically, a larger positive effect size implies a stronger association between the \textbf{first} {\color{red}attribute} in the attribute pair (\eg, ``male'') and the \textbf{first} {\color{blue}concept} in the concept pair (\eg, ``career'') AND a stronger association between the \textbf{second} {\color{red}attribute} in the attribute pair (\eg, ``female'') and the \textbf{second} {\color{blue}concept} in the concept pair (\eg, ``family''). A larger negative effect size, on the other hand, suggests a stronger association between the \textbf{first} {\color{red}attribute} (\eg, ``male'') and the \textbf{second} {\color{blue}concept} (\eg, ``family'') AND a stronger association between the \textbf{second} {\color{red}attribute} (\eg, ``female'') and the \textbf{first} {\color{blue}concept} (\eg, ``career''). From the results in Fig.~\ref{fig:eat}, we find that the social biases in image and text embeddings are not aligned but rather distributed very differently in terms of bias types. Specifically, in image embeddings, statistically significant biases related to gender-career are identified. In particular, the positive effect size observed in the Gender-Career test suggests the CLIP image encoder's tendency to associate the general concept of ``career'' with the attribute of ``male''. On the other hand, typical biases in text embeddings are related to gender-science and race.

\noindent \textbf{Summary.} One recent study, CLIP-clip~\cite{wang-etal-2021-gender}, addresses the social bias problem in both modalities. Nevertheless, it assumes the same level of biases in both modalities, which is, however, not well-grounded as observed in our probing experiments. For instance, the text bias related to gender-science is significant and has a negative effect size with a large magnitude, whereas its counterpart gender-science image bias is non-significant and has a positive effect size with a much smaller magnitude.

\newcommand{\veryshortarrow}[1][3pt]{\mathrel{%
   \hbox{\rule[\dimexpr\fontdimen22\textfont2-.2pt\relax]{#1}{.4pt}}%
   \mkern-4mu\hbox{\usefont{U}{lasy}{m}{n}\symbol{41}}}}
   
\section{Method} 
\subsection{Preliminary of CLIP}\label{sec:4.1}
CLIP aligns images and text via embedding them into the same latent space~\cite{Radford-2021-clip}. It employs a dual-stream architecture consisting of an image encoder and a text encoder. In particular, the sampled image and text are separately encoded into embeddings of the same dimension, followed by a cosine similarity estimation operation. During pre-training, CLIP utilises a contrastive loss to pull similar image and text embeddings together while pushing dissimilar ones away. For a sampled batch with $N$ matched text-image pairs, $\{(t_i, v_i)\}^N_{i=1}$, where $(t_i, v_i)$ denotes the $i$-th pair, the softmax-normalized text-to-image similarity score between the $j$-th text and the $k$-th image is defined as $s_t(t_j, v_k, \mathcal{V})$:
\vspace{-0.1cm}\begin{equation}\label{eq:1}
    s_t(t_j, v_k, \mathcal{V}) = \frac{\exp(\langle f(t_j), g(v_k) \rangle / \tau)}{\sum\limits_{v \in \mathcal{V}} \exp(\langle f(t_j), g(v) \rangle / \tau)},
\vspace{-0.1cm}\end{equation}
where $\mathcal{V}=\{v_i\}^N_{i=1}$, $f(.)$ and $g(.)$ denote the CLIP text and image encoder, respectively, $\tau$ is a learnable temperature parameter, and ${\langle \cdot , \cdot \rangle}$ represents the inner product between embeddings. The softmax-normalized image-to-text similarity scores between the $j$-th image and the $k$-th text, $s_v(t_k, v_j, \mathcal{T})$, is defined in a similar manner. CLIP maximises both $s_t(t_i, v_i, \mathcal{V})$ and $s_v(t_i, v_i, \mathcal{T})$ for the $i$-th matched pair, $(t_i, v_i)$.

\subsection{Dual-Bias Alignment Then Removal} \label{sec:4.2}
\noindent\textbf{Decoupling Bias Information. }Let $(t, v)$ denote a text-image pair associated with a specific social attribute (\eg, ``A photo of a female person''). Previous study~\cite{seth-2023-dear, chuang-2023-debiasing} shows their CLIP embeddings, such as $f(t)$, can be decomposed into a sum of bias and neutral information: \vspace{-0.2cm}\begin{equation}\label{eq:3} f(t) = \phi(t) + \overline{\phi}(t),
\vspace{-0.2cm}\end{equation}where $\phi(t)$ and $\overline{\phi}(t)$ represent the embedding of bias and neutral information in $t$, respectively. Under this context, the confounding bias factor $\phi(t)$ should describe the gender attribute - \textit{female}, where $\overline{\phi}(t)$ is enabled to be agnostic of any gender bias. For example, let $t'$ denote a counterpart text which only differs from $t$ by the gender bias direction (\eg, ``A photo of a \textit{male} person''), then the corresponding $\overline{\phi}(t')$ should be the same as $\overline{\phi}(t)$ as they represent the same neutral information. Similar notations can also be extended to the visual bias $\psi(v)$ and visual neutral information $\overline{\psi}(v)$.

\begin{figure*}[!tbp]
\vspace{-0.7cm}
\hspace{0.4cm}
    % \centering
    \includegraphics[width=2\columnwidth]{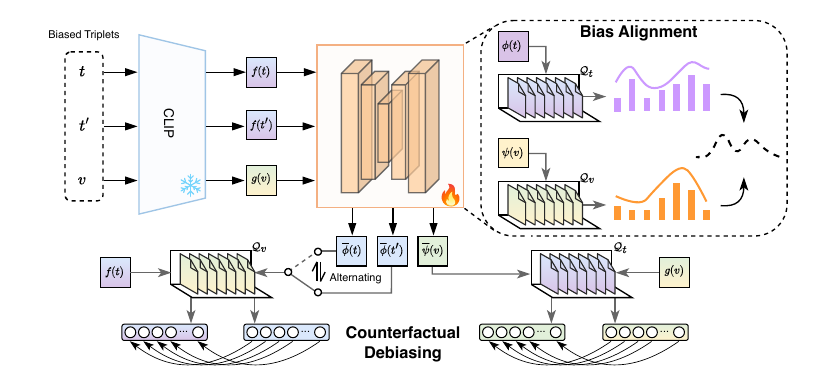}
    \vspace{-0.5cm}
    \caption{Overall pipeline. After obtaining the embedding of the given image, text, and counterfactual text using a frozen CLIP model, we first align the bias from both modalities with the help of two instantiated distributions. In addition, we design a counterfactual debiasing approach to bridge the embedding gap between two embeddings sharing the same concept yet with contrasting attributes.}
    % \vspace*{-0.5cm}
    \label{fig:method}
    \vspace{-0.6cm}
\end{figure*}

\noindent\textbf{Motivation.} Existing social bias removal methods for the CLIP model primarily suffer from the over-debiasing problem. The crux of it lies in the imbalanced treatment towards $\phi(t)$ and $\psi(v)$ in $(t, v)$. To mitigate this problem, we propose a novel dual-modal bias alignment-then-removal method. Specifically, our method involves two components: 1) image and text biases alignment; 2) dual-bias removal through counterfactual debiasing.

\noindent\textbf{Dual-Bias Alignment.} Considering the significant differences between image and text bias distributions (refer to Sect.~\ref{sec:3.2}), we propose a bias alignment module directly after CLIP encoders (refer to Fig.~\ref{fig:method}). We train this module while keeping the original CLIP model frozen. Specifically, we input the CLIP embeddings of a sampled batch of $N$ biased text-image pairs into the bias alignment module which outputs the aligned text and image bias embeddings: 
\begin{align}
    \phi(t_i)&\leftarrow \mathrm{BA}(f(t_i); \theta_\text{ba});\\
    \psi(v_i)&\leftarrow \mathrm{BA}(g(v_i); \theta_\text{ba}),
\end{align}
where $\mathrm{BA}(.; \theta_\text{ba})$ denotes our bias alignment module, parameterized by $\theta_\text{ba}$. After training, the aligned biases are expected to share similar bias distributions and can be removed without causing the over-debiasing problem, thus differentiating us from previous methods~\cite{chuang-2023-debiasing, wang-etal-2021-gender}. Subsequently, the debiased embeddings $\overline{\phi}(t_i)$ and $\overline{\psi}(v_i)$ can be obtained following Eqn.~\ref{eq:3}.

While our training goal is to align $\phi(t_i)$ and $\psi(v_i)$, directly enforcing element-wise matching via mean squared error or cosine similarity risks the loss of background information and feature diversity in image and text representations. Instead, we propose a novel approach to circumvent this problem. Our method is inspired by the moving queue mechanism in ALBEF~\cite{li-2021-albef} and MoCo~\cite{He-2020-moco}. Specifically, we maintain a text embedding queue, $\mathcal{Q}_t = \{f(t) \mid t\in\mathcal{T}_q\}$, and an image embedding queue, $\mathcal{Q}_v = \{g(v) \mid v\in\mathcal{V}_q\}$, where $\mathcal{T}_q=\{t_j\}^M_{j=1}$ and $\mathcal{V}_q=\{v_j\}^M_{j=1}$ are the most recently sampled $M$ texts and images, respectively, and $M$ is often much larger than $N$. Thereafter, we estimate the feature similarities between bias embeddings of each modality and its corresponding embedding queue as $p(t_i)$ and $p(v_i)$ using a scoring function $p(.)$, following Eqn.~\ref{eq:1} but in an intra-modal setup with a vectorized output. Specifically, $p(t_i)$ is expressed as,
\vspace{-0.1cm}
\hspace{-0.3cm}
\begin{equation}
    {\hspace{-0.3cm}p(t_i) = \left\{\frac{\exp(\langle \phi(t_i), f(t_l) \rangle / \tau)}{\sum_{m=1}^M \exp(\langle \phi(t_i), f(t_m) \rangle / \tau)}\right\}_{l=1}^{M}{\hspace{-0.6cm},}
\vspace{-0.1cm}{\hspace{-0.4cm}\tag{5}}}
\end{equation}where $t_l, t_m \in \mathcal{T}_q$ and $p(t_i)\in\mathbb{R}
^M$. The vector $p(v_i)$ is defined similarly.
The intuition behind this is that each pseudo distribution offers an intermediate proxy to view the given text or image bias from a global perspective. With the aid of these two distributions $p(t_i)$ and $p(v_i)$, we then design a bias alignment loss $\mathcal{L}_{ba}$ to align the bias between image and text by minimising the Kullback-Leibler (K-L) divergence of two distributions, 
% {\theta_{\text{ba}}}
\vspace{-0.0cm}\begin{equation}
\mathcal{L}_{\text{ba}} = \frac{1}{N}\sum_{i=1}^N D_\mathrm{KL}(p(t_i) \parallel p(v_i)),\vspace{-0.1cm}{\hspace{-0cm}\tag{6}}\end{equation}where $D_\mathrm{KL}$ denotes the K-L divergence loss function.

% \vspace{-0.1cm}
\begin{table*}[!t]
% \vspace{-0.5cm}
	\centering	
    \caption{\small{Gender and age debiasing results of three approaches trained on FairFace. ABLE is calculated based on in-domain fairness. We marked the best numbers pertaining to fairness results with \textbf{bold}, and those pertaining to the V-L alignment results on IN1K and Flickr with \underline{underline}. {\dag} implies equal or better V-L performance over the original CLIP.}}
    \vspace{-0.3cm}
    \small
        \scalebox{0.80}{
        \renewcommand{\arraystretch}{0.85}
        \begin{tabular}	{c | c | c |  c  c |c c | c  c | c c | c c | c}
        \toprule
        \multirow{3}{*}{Backbone} & \multirow{3}{*}{Biases} &\multirow{3}{*}{Methods} &  \multicolumn{2}{c|}{In-Domain} & \multicolumn{4}{c|}{Out-of-Domain} & \multicolumn{2}{c|}{\multirow{2}{*}{\makecell[c]{IN1K \\Acc. (\%)$\uparrow$}}} & \multicolumn{2}{c|}{\multirow{2}{*}{\makecell[c]{Flickr \\R@5 (\%)$\uparrow$}}} &\multirow{3}{*}{\makecell[c]{ABLE (\%)$\uparrow$}} \\
        \cline{4-9}& & & \multicolumn{2}{c|}{\scriptsize FairFace} & \multicolumn{2}{c|}{\scriptsize UTKFace} & \multicolumn{2}{c|}{\scriptsize FACET} & \multicolumn{2}{c|}{} & \multicolumn{2}{c|}{} \\
        & & &\scriptsize MS$\downarrow$& \scriptsize NDKL$\downarrow$ & \scriptsize MS$\downarrow$ & \scriptsize NDKL$\downarrow$ & \scriptsize MS$\downarrow$ & \scriptsize NDKL$\downarrow$ & \scriptsize Top-1 & \scriptsize Top-5 & \scriptsize TR & \scriptsize IR &  \\
        
        \midrule
        \multirow{8}{*}{\makecell[l]{ViT-B/16}} & \multirow{4}{*}{Gender} & \color{gray}Original CLIP & \color{gray}0.218 & \color{gray}0.088 & \color{gray}0.114 & \color{gray}0.080 & \color{gray}0.478 & \color{gray}0.215 & \color{gray}68.31 & \color{gray}91.83 & \color{gray}96.4 & \color{gray}85.5 & \color{gray}73.87 \\	
        & & CLIP-clip & 0.103 & 0.026 & 0.083 & 0.062 & 0.478 & 0.199 & 68.00 & 91.50 & 95.4 & 83.0 & 77.55 \\
        & & Biased-prompts & 0.161 & 0.048 & 0.179 & 0.062 & 0.460 & 0.215 & 65.07 & 89.38 & 94.3 & \underline{86.1}\dag & 73.78 \\
        & & Ours & \textbf{0.080} & \textbf{0.025} & \textbf{0.040} & \textbf{0.023} & \textbf{0.446} & \textbf{0.170} & \underline{68.05} & \underline{91.63} & \underline{96.6}\dag & 84.3 & \textbf{78.35} \\
        \cmidrule{2-14}& \multirow{4}{*}{Age} & \color{gray}Original CLIP & \color{gray}0.657 & \color{gray}0.433 & \color{gray}0.421 & \color{gray}0.229 & \color{gray}0.744 & \color{gray}0.367 & \color{gray}68.31 & \color{gray}91.83 & \color{gray}96.4 & \color{gray}85.5 & \color{gray}58.94 \\
        & & CLIP-clip & 0.647 & 0.432 & 0.402 & 0.215 & 0.742 & 0.373 & 67.97 & 91.61 & \underline{96.3} & 84.4 & 59.16 \\
        & & Biased-prompts & 0.777 & 0.550 & 0.578 & 0.451 & \textbf{0.635} & 0.355 & 66.43 & 90.28 & 94.1 & \underline{85.2} & 54.33 \\
        & & Ours & \textbf{0.608} & \textbf{0.294} & \textbf{0.377} & \textbf{0.115} & 0.738 & \textbf{0.341} & \underline{68.34}\dag & \underline{91.74} & 96.0 & 84.0 & \textbf{60.61} \\
        \midrule
        
        \multirow{8}{*}{\makecell[l]{ViT-B/32}} & \multirow{4}{*}{Gender} & \color{gray}Original CLIP & \color{gray}0.138 & \color{gray}0.054 & \color{gray}0.066 & \color{gray}0.032 & \color{gray}0.485 & \color{gray}0.225 & \color{gray}63.39 & \color{gray}88.83 & \color{gray}94.7 & \color{gray}83.5 & \color{gray}73.37 \\
        & & CLIP-clip & 0.107 & 0.030 & 0.061 & 0.023 & 0.492 & 0.215 & 59.62 & 86.29 & 90.9 & 76.2 & 71.68 \\
        & & Biased-prompts & 0.094 & \textbf{0.027} & 0.089 & 0.036 & \textbf{0.417} & \textbf{0.164} & 60.37 & 86.75 & 93.6 & 82.4 & 72.59 \\
        & & Ours & \textbf{0.090} & 0.030 & \textbf{0.050} & \textbf{0.021} & 0.466 & 0.204 & \underline{62.52} & \underline{88.56} & \underline{94.9}\dag & \underline{82.9} & \textbf{74.24} \\
        \cmidrule{2-14}& \multirow{4}{*}{Age} & \color{gray}Original CLIP & \color{gray}0.617 & \color{gray}0.416 & \color{gray}0.412 & \color{gray}0.253 & \color{gray}0.752 & \color{gray}0.388 & \color{gray}63.39 & \color{gray}88.83 & \color{gray}94.7 & \color{gray}83.5 & \color{gray}58.29 \\
        & & CLIP-clip & 0.635 & 0.425 & 0.400 & 0.252 & 0.749 & 0.387 & 62.40 & 88.30 & \underline{94.5} & 82.5 & 57.32 \\
        & & Biased-prompts & 0.701 & 0.497 & 0.522 & 0.409 & \textbf{0.663} & \textbf{0.366} & 61.07 & 86.92 & 92.0 & 82.2 & 54.76 \\
        & & Ours & \textbf{0.572} & \textbf{0.364} & \textbf{0.385} & \textbf{0.195} & 0.750 & 0.381 & \underline{63.13} & \underline{88.71} & 94.1 & \underline{82.8} & \textbf{59.60} \\

        \midrule
        \multirow{8}{*}{\makecell[l]{ViT-L/14}} & \multirow{4}{*}{Gender} & \color{gray}Original CLIP & \color{gray}0.185 & \color{gray}0.071 & \color{gray}0.066 & \color{gray}0.028 & \color{gray}0.487 & \color{gray}0.257 & \color{gray}75.55 & \color{gray}94.57 & \color{gray}97.2 & \color{gray}87.2 & \color{gray}79.14 \\
        & & CLIP-clip & 0.119 & 0.037 & \textbf{0.043} & \textbf{0.020} & 0.482 & 0.251 & 75.26 & 94.31 & \underline{96.9} & 85.1 & 81.47 \\
        & & Biased-prompts & 0.191 & 0.060 & 0.113 & 0.047 & 0.477 & \textbf{0.228} & 73.24 & 93.28 & 95.5 & \underline{87.2}\dag & 77.63 \\
        & & Ours & \textbf{0.106} & \textbf{0.035} & 0.069 & 0.026 & \textbf{0.475} & 0.239 & \underline{75.40} & \underline{94.53} & 96.8 & 87.0 & \textbf{82.04} \\
        \cmidrule{2-14}& \multirow{4}{*}{Age} & \color{gray}Original CLIP & \color{gray}0.672 & \color{gray}0.433 & \color{gray}0.619 & \color{gray}0.335 & \color{gray}0.749 & \color{gray}0.379 & \color{gray}75.55 & \color{gray}94.57 & \color{gray}97.2 & \color{gray}87.2 & \color{gray}60.96 \\
        & & CLIP-clip & 0.642 & 0.405 & 0.633 & 0.324 & 0.761 & 0.388 & 73.50 & 93.34 & 95.3 & 85.1 & 61.34 \\
        & & Biased-prompts & \textbf{0.564} & 0.412 & \textbf{0.433} & 0.299 & \textbf{0.727} & 0.393 & 74.00 & 93.35 & 94.5 & 86.6 & \textbf{64.34} \\
        & & Ours & 0.579 & \textbf{0.332} & 0.575 & \textbf{0.275} & 0.751 & \textbf{0.361} & \underline{75.26} & \underline{94.46} & \underline{97.1} & \underline{86.8} & 64.25 \\
        
        \midrule
        \multirow{8}{*}{\makecell[l]{ViT-H/14}} & \multirow{4}{*}{Gender} & \color{gray}Original CLIP & \color{gray}0.193 & \color{gray}0.079 & \color{gray}0.127 & \color{gray}0.056 & \color{gray}0.483 & \color{gray}0.250 & \color{gray}77.95 & \color{gray}95.19 & \color{gray}99.5 & \color{gray}94.1 & \color{gray}80.14 \\
        & & CLIP-clip & 0.139 & \textbf{0.049} & \textbf{0.087} & \textbf{0.032} & 0.465 & \textbf{0.211} & 76.35 & 94.29 & 97.8 & 91.6 & 81.34 \\
        & & Biased-prompts & 0.172 & 0.055 & 0.110 & 0.043 & \textbf{0.456} & 0.217 & 77.53 & 95.00 & 99.2 & 93.3 & 80.62 \\
        & & Ours & \textbf{0.138} & 0.051 & 0.100 & 0.040 & 0.466 & 0.233 & \underline{77.67} & \underline{95.12} & \underline{99.6}\dag & \underline{94.1}\dag & \textbf{82.11} \\
        \cmidrule{2-14}& \multirow{4}{*}{Age} & \color{gray}Original CLIP & \color{gray}0.569 & \color{gray}0.368 & \color{gray}0.498 & \color{gray}0.239 & \color{gray}0.764 & \color{gray}0.390 & \color{gray}77.95 & \color{gray}95.19 & \color{gray}99.5 & \color{gray}94.1 & \color{gray}65.59 \\
        & & CLIP-clip & \textbf{0.482} & 0.356 & 0.432 & 0.214 & 0.796 & 0.437 & 76.67 & 94.65 & 98.3 & 92.2 & \textbf{68.41} \\
        & & Biased-prompts & 0.564 & 0.387 & 0.452 & 0.264 & 0.790 & 0.433 & 77.11 & 94.94 & 98.8 & 93.0 & 65.40 \\
        & & Ours & 0.515 & \textbf{0.289} & \textbf{0.409} & \textbf{0.141} & \textbf{0.758} & \textbf{0.377} & \underline{77.85} & \underline{95.18} & \underline{99.4} & \underline{94.0} & 67.62 \\
        \bottomrule	
        \end{tabular}}
	\label{table:debiasing_gender_fairface}\vspace{-0.4cm}
\end{table*}

\noindent\textbf{Conterfactual Debiasing.} Complementary to $\mathcal{L}_{ba}$, we propose an additional counterfactual debiasing objective to supervise dual-modal debiasing. For each $t_i$, we obtain its counterfactual embedding $t_i'$ by replacing the attribute keyword in the original text prompt. Both $\overline{\phi}(t_i)$ and $\overline{\phi}(t_i')$ are expected to express the same neutral concept if debiased correctly. Our counterfactual debiasing objective is designed to pull $\overline{\phi}(t_i)$ and $\overline{\phi}(t_i')$ closer in the embedding space while maintaining their V-L alignment capabilities as those of $f(t)$. The text counterfactual debiasing loss is expressed as a cross-entropy loss between text-to-image similarities of the debiased embeddings and the original embeddings:
\vspace{-0.2cm}\begin{equation}
\hspace{-0.0cm}\mathcal{L}_{cd}^t =  -\frac{1}{N}\sum_{i=1}^N \left[\sum\limits_{v\in\mathcal{V}_q}s_t(t_i, v, \mathcal{V}_q)\log{\overline{s}_t(a(t_i, t_i'), v, \mathcal{V}_q)}\right]{\hspace{-0.18cm},}
\vspace{-0.0cm}{\hspace{0.0cm}\tag{7}}
\end{equation}
where $a(t_i, t_i') = \beta t_i + (1-\beta)t_i'$ is an alternating function with $\beta\sim\text{Bernoulli}(0.5)$, $s_t(.,.,.)$ and $\overline{s}_t(.,.,.)$ denote the text-to-image similarity of original text embedding and debiased text embedding, respectively. We calculate $s_t(.,.,.)$ following Eqn.~\ref{eq:1} and define $\overline{s}_t(.,.,.)$ similarly as follows:
\vspace{-0.1cm}\begin{equation}\hspace{-1.5cm}
    \overline{s}_t(t_j, v_k, \mathcal{V}_q) = \frac{\exp(\langle \overline{\phi}(t_j), g(v_k) \rangle / \tau)}{\sum\limits_{v\in\mathcal{V}_q} \exp(\langle \overline{\phi}(t_j), g(v) \rangle / \tau)}.
\vspace{-0.1cm}{\hspace{-1.2cm}\tag{8}}\end{equation}
We leverage $\mathcal{L}^t_{cd}$ to i) inherit the original V-L alignment capability, and ii) bridge the embedding gap of two text embeddings with the same concept yet different attributes. In this way, the social bias between different attributes can be accordingly alleviated. 

For image embeddings, a similar loss is applied to align the image bias removal with that of text and subsequently maintain the V-L alignment of the debiased image embeddings with text embeddings: 

\vspace{0.0cm}\begin{equation}
\hspace{-1.5cm}\mathcal{L}_{cd}^v =  -\frac{1}{N}\sum_{i=1}^N \left[\sum\limits_{t\in\mathcal{T}_q}s_v(t, v_i,\mathcal{T}_q)\log{\overline{s}_v(t, v_i,\mathcal{T}_q)}\right]{\hspace{-0.1cm},}
\vspace{-0.0cm}{\hspace{-1.2cm}\tag{9}}
\end{equation}which does not involve counterfactual images due to their inaccessibility\footnote{Although counterfactual image embeddings can be created using generative models, we abandon such an operation due to computing resource and generation unfaithfulness reasons.}. Combining both components, the final counterfactual debiasing loss is $\mathcal{L}_{cd} = (\mathcal{L}_{cd}^t + \mathcal{L}_{cd}^v)/2$.

\subsection{Overall Training and Inference} 
For training, our two training objectives are combined as $\mathcal{L} = \alpha\mathcal{L}_{cd} + (1-\alpha)\mathcal{L}_{ba}$ with a hyperparameter $\alpha \in [0, 1]$. During inference in downstream tasks, our optimized $\mathrm{BA}(.; \theta_\text{ba})$ can be used as a plug-and-play module in CLIP and produce the debiased embeddings $(\overline{\phi}(t), \overline{\psi}(v))$ of $(t, v)$, respectively expressed as: 
\begin{align}
    \overline{\phi}(t)&\leftarrow f(t) - \mathrm{BA}(f(t); \theta_\text{ba});{\tag{10}}\\
    \overline{\psi}(v)&\leftarrow g(v) - \mathrm{BA}(g(v); \theta_\text{ba}).{\tag{11}}
\end{align}

\section{Experimental Settings}
\noindent\textbf{Datasets.} For model training, we utilised FairFace~\cite{kar-2021-fairface} and UTKFace~\cite{zhang-2017-utkface} training subsets containing human face images with gender, age and race labels. Unlike previous studies, for a more comprehensive evaluation, we employed both in-domain and out-of-domain fairness evaluations by interchangeably using another test set for the current trained model evaluation. A recent dataset FACET~\cite{gusta-2023-facet} was also used to introduce additional out-of-domain photos of full-body images. Through these datasets, we evaluated both the fairness and generalizability of debiased models. Moreover, to evaluate the V-L task performance of debiased CLIP, we used the ImageNet-1K~\cite{deng-2009-imagenet} and Flickr-1K~\cite{plummer-2015-flickr} for classification and retrieval tasks, respectively.

\begin{table*}[!ht]
\vspace{-0.4cm}
% \tablestyle{2.5pt}{1.0}
	\centering	
    \caption{\small{Gender and age debiasing results of three approaches trained on UTKFace.}}
    \vspace{-0.3cm}
    \small
        \scalebox{0.80}{
        \renewcommand{\arraystretch}{0.85}
        \begin{tabular}	{c | c | c |  c  c |c c | c  c | c c | c c | c}
        \toprule
        \multirow{3}{*}{Backbone} & \multirow{3}{*}{Biases} &\multirow{3}{*}{Methods} &  \multicolumn{2}{c|}{In-Domain} & \multicolumn{4}{c|}{Out-of-Domain} & \multicolumn{2}{c|}{\multirow{2}{*}{\makecell[c]{IN1K \\Acc. (\%)$\uparrow$}}} & \multicolumn{2}{c|}{\multirow{2}{*}{\makecell[c]{Flickr \\R@5 (\%)$\uparrow$}}} &\multirow{3}{*}{\makecell[c]{ABLE (\%)$\uparrow$}} \\
        \cline{4-9}& & & \multicolumn{2}{c|}{\scriptsize UTKFace} & \multicolumn{2}{c|}{\scriptsize Fairface} & \multicolumn{2}{c|}{\scriptsize FACET} & \multicolumn{2}{c|}{} & \multicolumn{2}{c|}{} \\
        & & &\scriptsize MS$\downarrow$ & \scriptsize NDKL$\downarrow$ & \scriptsize MS$\downarrow$ & \scriptsize NDKL$\downarrow$ & \scriptsize MS$\downarrow$ & \scriptsize NDKL$\downarrow$ & \scriptsize Top-1 & \scriptsize Top-5 & \scriptsize TR & \scriptsize IR &  \\       
        \midrule
        \multirow{8}{*}{\makecell[l]{ViT-B/16}} & \multirow{4}{*}{Gender} & \color{gray}Original CLIP & \color{gray}0.114 & \color{gray}0.080 & \color{gray}0.218 & \color{gray}0.088 & \color{gray}0.478 & \color{gray}0.215 & \color{gray}68.31 & \color{gray}91.83 & \color{gray}96.4 & \color{gray}85.5 & \color{gray}77.39 \\	
        & & CLIP-clip & 0.070 & 0.055 & 0.133 & 0.038 & 0.459 & 0.190 & 67.81 & 91.42 & 95.4 & 83.0 & 78.52 \\
        & & Biased-prompts & 0.179 & 0.062 & 0.161 & 0.048 & 0.460 & 0.215 & 65.07 & 89.38 & 94.3 & \underline{86.1}\dag & 73.18 \\
        & & Ours & \textbf{0.048} & \textbf{0.043} & \textbf{0.101} & \textbf{0.032} & \textbf{0.456} & \textbf{0.181} & \underline{67.99} & \underline{91.64} & \underline{95.8} & 84.6 & \textbf{79.36} \\
        \cmidrule{2-14}& \multirow{4}{*}{Age} & \color{gray}Original CLIP & \color{gray}0.421 & \color{gray}0.229 & \color{gray}0.657 & \color{gray}0.433 & \color{gray}0.744 & \color{gray}0.367 & \color{gray}68.31 & \color{gray}91.83 & \color{gray}96.4 & \color{gray}85.5 & \color{gray}66.96 \\
        & & CLIP-clip & \textbf{0.393} & \textbf{0.215} & 0.643 & 0.430 & 0.745 & 0.364 & \underline{67.93} & \underline{91.58} & \underline{96.1} & 84.5 & \textbf{67.70} \\
        & & Biased-prompts & 0.578 & 0.451 & 0.777 & 0.550 & \textbf{0.635} & \textbf{0.355} & 66.43 & 90.28 & 94.1 & \underline{85.2} & 60.83 \\
        & & Ours & 0.414 & 0.231 & \textbf{0.606} & \textbf{0.410} & 0.746 & 0.365 & 67.63 & 91.46 & 95.6 & 84.6 & 66.86 \\
        \midrule
        \multirow{8}{*}{\makecell[l]{ViT-B/32}} & \multirow{4}{*}{Gender} & \color{gray}Original CLIP & \color{gray}0.066 & \color{gray}0.032 & \color{gray}0.138 & \color{gray}0.054 & \color{gray}0.485 & \color{gray}0.225 & \color{gray}63.39 & \color{gray}88.83 & \color{gray}94.7 & \color{gray}83.5 & \color{gray}75.60 \\
        & & CLIP-clip & 0.098 & 0.045 & 0.253 & 0.105 & 0.500 & 0.240 & 62.21 & \underline{88.23} & 93.0 & 81.0 & 73.79 \\
        & & Biased-prompts & 0.089 & 0.036 & \textbf{0.094} & \textbf{0.027} & \textbf{0.417} & \textbf{0.164} & 60.37 & 86.75 & 93.6 & 82.4 & 72.74 \\
        & & Ours & \textbf{0.043} & \textbf{0.033} & 0.108 & 0.039 & 0.469 & 0.212 & \underline{62.46} & \underline{88.23} & \underline{94.7}\dag & \underline{82.9} & \textbf{75.60} \\
        \cmidrule{2-14}& \multirow{4}{*}{Age} & \color{gray}Original CLIP & \color{gray}0.412 & \color{gray}0.253 & \color{gray}0.617 & \color{gray}0.416 & \color{gray}0.752 & \color{gray}0.388 & \color{gray}63.39 & \color{gray}88.83 & \color{gray}94.7 & \color{gray}83.5 & \color{gray}64.77 \\
        & & CLIP-clip & 0.415 & 0.264 & 0.659 & 0.435 & 0.754 & 0.397 & 62.70 & 88.31 & \underline{94.2} & \underline{83.2} & 64.34 \\
        & & Biased-prompts & 0.522 & 0.409 & 0.701 & 0.497 & \textbf{0.663} & \textbf{0.366} & 61.07 & 86.92 & 92.0 & 82.2 & 60.19 \\
        & & Ours & \textbf{0.407} & \textbf{0.252} & \textbf{0.627} & \textbf{0.416} & 0.751 & 0.370 & \underline{62.93} & \underline{88.66} & 94.1 & 82.5 & \textbf{64.69} \\

        \midrule
        
        \multirow{8}{*}{\makecell[l]{ViT-L/14}} & \multirow{4}{*}{Gender} & \color{gray}Original CLIP & \color{gray}0.066 & \color{gray}0.028 & \color{gray}0.185 & \color{gray}0.071 &\color{gray}0.487 & \color{gray}0.257 & \color{gray}75.55 & \color{gray}94.57 & \color{gray}97.2 & \color{gray}87.2 & \color{gray}83.61 \\
        & & CLIP-clip & 0.059 & \textbf{0.021} & \textbf{0.097} & \textbf{0.030} & \textbf{0.465} & \textbf{0.223} & 75.23 & 94.37 & 96.6 & 85.2 & 83.68 \\
        & & Biased-prompts & 0.113 & 0.047 & 0.191 & 0.060 & 0.477 & 0.228 & 73.24 & 93.28 & 95.5 & \underline{87.2}\dag & 80.48 \\
        & & Ours & \textbf{0.057} & 0.026 & 0.175 & 0.064 & 0.484 & 0.253 & \underline{75.41} & \underline{94.53} & \underline{97.0} & 87.0 & \textbf{83.85} \\
        \cmidrule{2-14}& \multirow{4}{*}{Age} & \color{gray}Original CLIP & \color{gray}0.619 & \color{gray}0.335 & \color{gray}0.672 & \color{gray}0.433 & \color{gray}0.749 & \color{gray}0.379 & \color{gray}75.55 & \color{gray}94.57 & \color{gray}97.2 & \color{gray}87.2 & \color{gray}62.87 \\
        & & CLIP-clip & 0.666 & 0.384 & 0.679 & 0.432 & 0.753 & 0.386 & 75.31 & 94.44 & 96.2 & 86.4 & 61.08 \\
        & & Biased-prompts & \textbf{0.433} & 0.299 & \textbf{0.564} & 0.412 & \textbf{0.727} & 0.393 & 74.00 & 93.35 & 94.5 & 86.6 & \textbf{69.11} \\
        & & Ours & 0.577 & \textbf{0.270} & 0.568 & \textbf{0.332} & 0.754 & \textbf{0.362} & \underline{75.32} & \underline{94.54} & \underline{97.3}\dag & \underline{86.8} & 64.33 \\
        \midrule
        \multirow{8}{*}{\makecell[l]{ViT-H/14}} & \multirow{4}{*}{Gender} & \color{gray}Original CLIP & \color{gray}0.127 & \color{gray}0.056 & \color{gray}0.193 & \color{gray}0.079 &\color{gray}0.483 & \color{gray}0.250 & \color{gray}77.95 & \color{gray}95.19 & \color{gray}99.5 & \color{gray}94.1 & \color{gray}82.70 \\
        & & CLIP-clip & \textbf{0.075} & \textbf{0.026} & \textbf{0.135} & \textbf{0.045} & \textbf{0.450} & \textbf{0.191} & 77.57 & 94.90 & 98.4 & 93.0 & \textbf{84.50} \\
        & & Biased-prompts & 0.110 & 0.043 & 0.172 & 0.055 & 0.456 & 0.217 & 77.53 & 95.00 & 99.2 & 93.3 & 82.44 \\
        & & Ours & 0.110 & 0.044 & 0.155 & 0.056 & 0.455 & 0.217 & \underline{77.84} & \underline{95.17} & \underline{99.5}\dag & \underline{93.8} & 83.30 \\
        \cmidrule{2-14}& \multirow{4}{*}{Age} & \color{gray}Original CLIP & \color{gray}0.498 & \color{gray}0.239 & \color{gray}0.569 & \color{gray}0.368 & \color{gray}0.764 & \color{gray}0.390 & \color{gray}77.95 & \color{gray}95.19 & \color{gray}99.5 & \color{gray}94.1 & \color{gray}68.30 \\
        & & CLIP-clip & 0.457 & \textbf{0.209} & \textbf{0.519} & \textbf{0.340} & 0.769 & \textbf{0.385} & 77.50 & 95.03 & 99.2 & 93.8 & 69.69 \\
        & & Biased-prompts & \textbf{0.452} & 0.264 & 0.564 & 0.387 & 0.790 & 0.433 & 77.11 & 94.94 & 98.8 & 93.0 & \textbf{69.73} \\
        & & Ours & 0.473 & 0.219 & 0.565 & 0.366 & \textbf{0.766} & 0.399 & \underline{77.82} & \underline{95.19}\dag & \underline{99.5}\dag & \underline{94.0} & 69.21 \\
        \bottomrule	
        \end{tabular}}
	\label{table:debiasing_gender_utkface}\vspace{-0.4cm}
\end{table*}

\noindent\textbf{Fairness Metrics.}
Following existing work~\cite{berg-etal-2022-prompt,seth-2023-dear}, we leveraged retrieval-based metrics, mean $\text{MaxSkew}@k$ and mean $\text{NDKL}@k$, which respectively represent the largest unfair advantage and the average unfair advantage given to images with a certain attribute in the retrieval task. For both metrics, a smaller value implies more fairness of the model.

\noindent\textbf{V-L Alignment Metrics.}
We report the zero-shot performance on ImageNet-1k classification task~\cite{deng-2009-imagenet} (Top-1 and Top-5 accuracies) and the Flickr-1k retrieval tasks~\cite{plummer-2015-flickr} (Recall@5 scores for image-to-text retrieval (TR) and text-to-image retrieval (IR)).

\noindent\textbf{Alignment and Bias Level Evaluation (ABLE).} The aforementioned evaluation metrics either evaluate the level of fairness or the V-L alignment of debiasing methods. However, we care about the holistic assessment of social bias removal that covers the above two aspects. Inspired by the F1 Score for imbalanced classification problems, we advocate an Alignment and Bias Level Evaluation (ABLE) metric, defined as,
\vspace{-0.2cm}
\begin{equation}{
\mathrm{ABLE} = \frac{2}{\frac{1}{acc} + \frac{1}{\exp(-\text{MaxSkew}@k)}},\vspace{-1.5cm}}{\tag{12}}\vspace{-0.2cm}
\end{equation}
which is the harmonic mean of the zero-shot ImageNet Top-1 accuracy $acc$ and the negative exponential of $\text{MaxSkew}@k$, representing the V-L alignment and fairness level, respectively. Its range is within [0, 1], and a larger value denotes a better model.

\noindent\textbf{Compared Baselines}
We compared our approach with the original undebiased CLIP and two other publicly available debiasing baselines, CLIP-clip~\cite{wang-etal-2021-gender} and Biased-prompts~\cite{chuang-2023-debiasing} on ViT-B/16, ViT-B/32, ViT-L/14 and ViT-H/14 backbones, respectively. \textbf{CLIP-clip} relies on the training dataset to estimate the mutual information between image embedding dimensions and the attribute labels. It then removes the most biased dimensions with the largest amount of mutual information. \textbf{Biased-prompts} is training-free and calculates a calibrated projection matrix based on biased text prompts to debias text embeddings. All methods were trained to remove either one of the three common types of social biases: gender, age and race.
\section{Experimental Results}
\subsection{Overall Debiasing Results}
The results for gender and age debiasing are shown in Table~\ref{table:debiasing_gender_fairface} and Table~\ref{table:debiasing_gender_utkface}. Kindly refer to the supplementary material for more results. We make the following observations:
\begin{itemize}[leftmargin=8pt]
\setlength\itemsep{0em}
    \item Previous debiasing methods such as CLIP-clip~\cite{wang-etal-2021-gender} and Biased-prompts~\cite{chuang-2023-debiasing} achieve significant bias reduction but lead to a drastic drop in V-L zero-shot performance, \ie, with a lower ABLE score. In contrast, our method obtains the best ABLE across most settings, suggesting that it achieves a better trade-off between debiasing and V-L alignment. Specifically, our method exhibits comparable or even better debiasing capabilities in both in-domain and out-of-domain fairness evaluations, showcasing its superior generalizability.
    \item In terms of V-L alignment, our method preserves the original CLIP's zero-shot performance with less than 1 point of drop in classification accuracy and retrieval R@5 scores. It sometimes even demonstrates equal or better V-L performance than the original CLIP model.
\end{itemize}

\begin{table}[!h]
% \hspace{-1cm}
% \tablestyle{3pt}{1.0}
	\centering	
    \caption{\small{Ablation study results for gender debiasing with the ViT-B/16 backbone trained on FairFace.}}
    \vspace{-0.8em}
    \small
    \scalebox{0.81}{
        \hspace{-0.27cm}
        \setlength{\tabcolsep}{2.6pt}
        \begin{tabular}	{c | c  c | c  c | c c | c c | c}
        \toprule
        \multirow{2}{*}{Methods} &  \multicolumn{2}{c|}{FairFace} & \multicolumn{2}{c|}{FACET} & \multicolumn{2}{c|}{IN1K (\%)$\uparrow$} & \multicolumn{2}{c|}{Flickr (\%)$\uparrow$} & ABLE \\
        & \scriptsize MS$\downarrow$ & \scriptsize NDKL$\downarrow$ & \scriptsize MS$\downarrow$ & \scriptsize NDKL$\downarrow$ & \scriptsize Top-1 & \scriptsize Top-5 & \scriptsize TR & \scriptsize IR & (\%)$\uparrow$ \\
        \midrule
        \color{gray}Undebiased & \color{gray}0.218 & \color{gray}0.088 & \color{gray}0.478 & \color{gray}0.215 & \color{gray}68.31 & \color{gray}91.83 & \color{gray}96.4 & \color{gray}85.5 & \color{gray}73.87 \\
        Ours (complete) & \textbf{0.080} & \textbf{0.025} & 0.446 & 0.170 & 68.05 & 91.63 & \underline{96.6}\dag & 84.3 & \textbf{78.35} \\
        Ours (w/o $\mathcal{L}_{cd})$ & 0.167 & 0.056 & 0.467 & 0.199 & \underline{68.28} & \underline{91.80} & 96.3 & \underline{84.8} & 75.58 \\
        Ours (w/o $\mathcal{L}_{ba})$ & 0.095 & 0.033 & \textbf{0.436} & \textbf{0.158} & 67.84 & 91.65 & 96.2 & 83.4 & 77.71 \\
        \bottomrule	
        \end{tabular}}
	\label{table:ablation_study_gender}
 % \vspace*{-0.5cm}
\end{table}

\begin{table}[!h]
\vspace{-0.3cm}
% \hspace{-1cm}
% \tablestyle{3pt}{1.0}
	\centering	
    \caption{\small{Ablation study results for age debiasing.}}
    \vspace{-0.8em}
    \small
    \scalebox{0.81}{
        \hspace{-0.35cm}
        \setlength{\tabcolsep}{2.7pt}
        \begin{tabular}	{c | c  c | c  c | c c | c c | c}
        \toprule
        \multirow{2}{*}{Methods} &  \multicolumn{2}{c|}{FairFace} & \multicolumn{2}{c|}{FACET} & \multicolumn{2}{c|}{IN1K (\%)$\uparrow$} & \multicolumn{2}{c|}{Flickr (\%)$\uparrow$} & ABLE \\
        & \scriptsize MS$\downarrow$ & \scriptsize NDKL$\downarrow$ & \scriptsize MS$\downarrow$ & \scriptsize NDKL$\downarrow$ & \scriptsize Top-1 & \scriptsize Top-5 & \scriptsize TR & \scriptsize IR & (\%)$\uparrow$ \\
        \midrule
        \color{gray}Undebiased & \color{gray}0.657 & \color{gray}0.433 & \color{gray}0.744 & \color{gray}0.367 & \color{gray}68.31 & \color{gray}91.83 & \color{gray}96.4 & \color{gray}85.5 & \color{gray}58.94 \\
        Ours (complete) & \textbf{0.608} & \textbf{0.294} & 0.738 & \textbf{0.341} & 68.34\dag & 91.74 & 96.0 & 84.0 & \textbf{60.61} \\
        Ours (w/o $\mathcal{L}_{cd})$ & 0.642 & 0.430 & 0.745 & 0.363 & \underline{68.36}\dag & \underline{91.79} & \underline{96.3} & \underline{85.2} & 59.47 \\
        Ours (w/o $\mathcal{L}_{ba})$ & 0.691 & 0.463 & \textbf{0.737} & 0.362 & 67.90 & 91.59 & 96.0 & 84.2 & 56.14 \\
        \bottomrule	
        \end{tabular}}
	\label{table:ablation_study_age}
 % \vspace*{-0.5cm}
\end{table}

\begin{table}[!h]
\vspace{-0.3cm}
% \hspace{-1cm}
% \tablestyle{3pt}{1.0}
	\centering	
    \caption{\small{Ablation study results for race debiasing.}}
    \vspace{-0.8em}
    \small
    \scalebox{0.81}{
        \hspace{-0.41cm}
        \setlength{\tabcolsep}{2.7pt}
        \begin{tabular}	{c | c  c | c  c | c c | c c | c}
        \toprule
        \multirow{2}{*}{Methods} &  \multicolumn{2}{c|}{FairFace} & \multicolumn{2}{c|}{UTKFace} & \multicolumn{2}{c|}{IN1K (\%)$\uparrow$} & \multicolumn{2}{c|}{Flickr (\%)$\uparrow$} & ABLE \\
        & \scriptsize MS$\downarrow$ & \scriptsize NDKL$\downarrow$ & \scriptsize MS$\downarrow$ & \scriptsize NDKL$\downarrow$ & \scriptsize Top-1 & \scriptsize Top-5 & \scriptsize TR & \scriptsize IR & (\%)$\uparrow$ \\
        \midrule
        \color{gray}Undebiased & \color{gray}0.528 & \color{gray}0.182 & \color{gray}0.575 & \color{gray}0.137 & \color{gray}68.31 & \color{gray}91.83 & \color{gray}96.4 & \color{gray}85.5 & \color{gray}63.31 \\
        Ours (complete) & \textbf{0.353} & \textbf{0.125} & \textbf{0.378} & \textbf{0.069} & 68.07 & 91.64 & \underline{96.5}\dag & 83.8 & \textbf{69.14} \\
        Ours (w/o $\mathcal{L}_{cd})$ & 0.540 & 0.184 & 0.591 & 0.147 & \underline{68.34}\dag & \underline{91.84}\dag & 96.4\dag & \underline{85.3} & 62.91 \\
        Ours (w/o $\mathcal{L}_{ba})$ & 0.492 & 0.159 & 0.445 & 0.087 & 67.83 & 91.67 & 95.9 & 84.1 & 64.31 \\
        \bottomrule	
        \end{tabular}}
	\label{table:ablation_study_race}
 % \vspace*{-0.5cm}
\end{table}

\begin{figure}[!tbp]
% \vspace{-0.7cm}
\hspace{-0.32cm}
    % \centering
    \includegraphics[width=1.05\columnwidth]{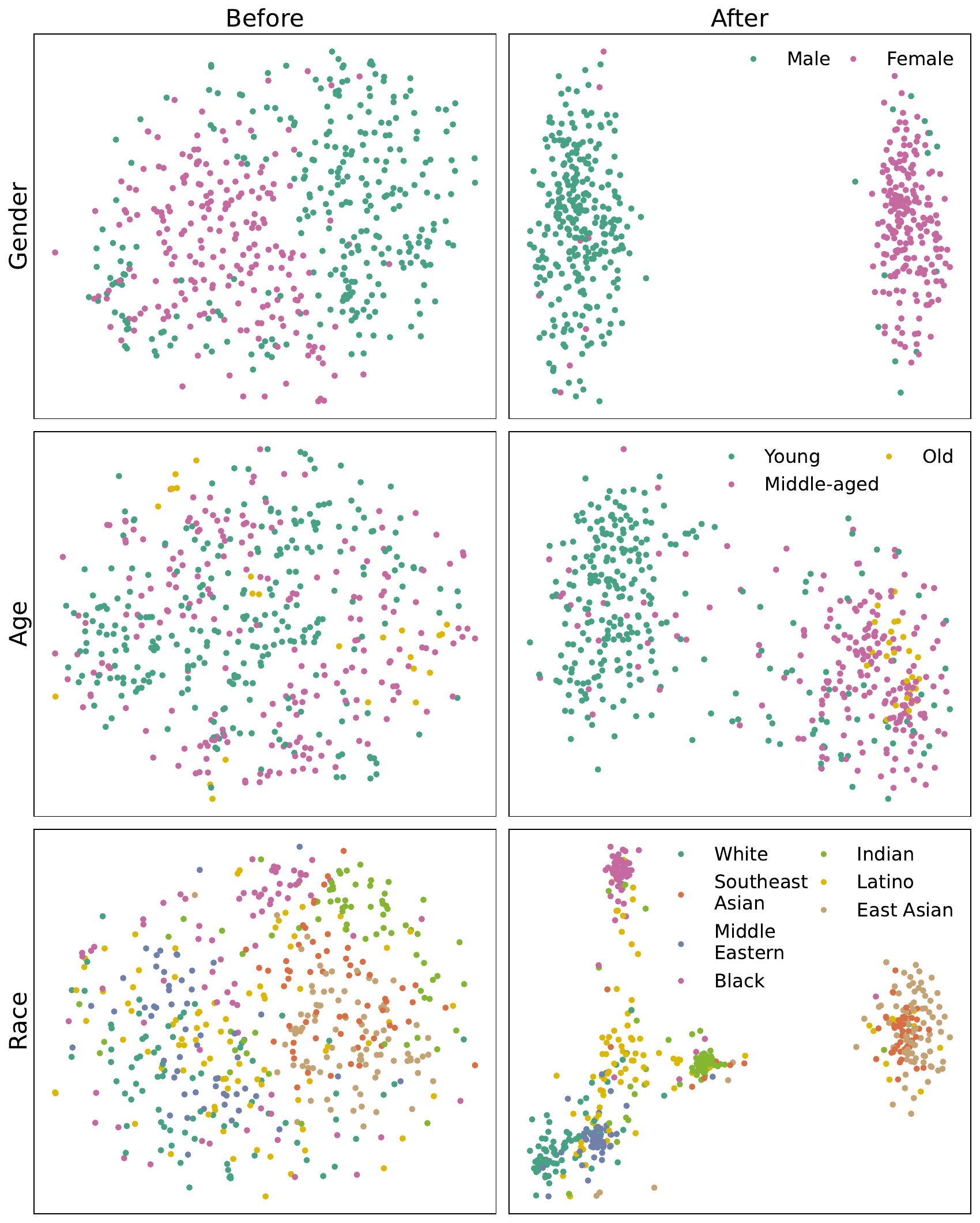}
    % \vspace{-1.4cm}
    \caption[width=1\columnwidth]{T-SNE plot of the bias information $\psi(v_i)$ in sampled $v_i$, estimated by our bias alignment module before and after training.}
    % \vspace*{-0.5cm}
    \label{fig:vis_results}
    \vspace{-0.6cm}
\end{figure}

\begin{table}[!h]
% \vspace{-0.5cm}
% \tablestyle{2.5pt}{1.0}
	\centering	
    \caption{\small{Universal debiasing results with the ViT-B/16 backbone trained on FairFace. - represents unavailable race labels.}}
    \vspace{-0.8em}
    \small
        \scalebox{0.81}{
        \hspace{-0.35cm}
        \setlength{\tabcolsep}{2.2pt}
        \begin{tabular}	{c | c  c | c  c | c c | c c | c}
        \toprule
        \multirow{2}{*}{Methods} & \multicolumn{2}{c|}{FairFace} & \multicolumn{2}{c|}{FACET} & \multicolumn{2}{c|}{IN1K (\%)$\uparrow$} & \multicolumn{2}{c|}{Flickr (\%)$\uparrow$} & ABLE\\
        & \scriptsize MS$\downarrow$ & \scriptsize NDKL$\downarrow$ & \scriptsize MS$\downarrow$ & \scriptsize NDKL$\downarrow$ & \scriptsize Top-1 & \scriptsize Top-5 & \scriptsize TR & \scriptsize IR & (\%)$\uparrow$ \\
        \midrule
        \color{gray}Original (gender) & \color{gray}0.218 & \color{gray}0.088 & \color{gray}0.478 & \color{gray}0.215 & \multirow{3}{*}{\makecell[l]{\color{gray}68.31}} & \multirow{3}{*}{\makecell[l]{\color{gray}91.83}} & \multirow{3}{*}{\makecell[l]{\color{gray}96.4}} & \multirow{3}{*}{\makecell[l]{\color{gray}85.5}} & \color{gray}73.87 \\
         \color{gray}Original (age) & \color{gray}0.657 & \color{gray}0.433 & \color{gray}0.744 & \color{gray}0.367 & & & & & \color{gray}58.94 \\
         \color{gray}Original (race) & \color{gray}0.528 & \color{gray}0.182 & - & - & & & & & \color{gray}63.31 \\
        \midrule
        Ours (gender) & \textbf{0.101} & \textbf{0.032} & \textbf{0.443} & \textbf{0.168} & \multirow{3}{*}{\makecell[l]{68.32\dag}} & \multirow{3}{*}{\makecell[l]{91.68}} & \multirow{3}{*}{\makecell[l]{96.1}} & \multirow{3}{*}{\makecell[l]{83.8}} & \textbf{77.82} \\
        Ours (age) & \textbf{0.649} & \textbf{0.425} & \textbf{0.730} & \textbf{0.347} & & & & & \textbf{59.22} \\
        Ours (race) & \textbf{0.471} & \textbf{0.142} & - & - & & & & & \textbf{65.25} \\
        \bottomrule	
        \end{tabular}}
	\label{table:joint_debiasing}
 \vspace*{-0.55cm}
\end{table}

\subsection{Visualization of Removed Bias}
To provide more evidence of bias removal beyond quantitative results, we use t-SNE to visualize the visual bias embedding  $\psi(v_i)$ corresponding to $v_i$ in the training batch before and after module training. We observe in Fig.~\ref{fig:vis_results} that after training with our bias alignment and counterfactual debiasing objectives, the image bias embeddings removed from the original embeddings carry significant bias information and can be clustered according to specific social groups of gender, age and race.

\subsection{Ablation Study}
We conduct detailed ablation studies under various debiasing setups and show the results in Table~\ref{table:ablation_study_gender}, Table~\ref{table:ablation_study_age} and Table~\ref{table:ablation_study_race}. We observe that our method without the counterfactual debiasing loss leads to a great degradation in performance, especially for debiasing. This suggests that the counterfactual debiasing objective is critical in guiding the removal of image and text biases. Second, our method without the bias alignment operation also suffers a drop in debiasing together with a significant degradation of V-L alignment. This further highlights the necessity of aligning image and text biases before debiasing. Combining both losses delivers an improved balance of debiasing and V-L performance with the highest ABLE score.

\subsection{Universal Bias Removal}
We also study whether our method can be applied for reducing multiple types of social biases at the same time, which is not explored in previous studies such as CLIP-clip and Biased-prompts. The debiasing is conducted by varying all attributes in the biased text prompts simultaneously during training. The results are shown in Table~\ref{table:joint_debiasing}. Our model has achieved the joint removal of all three types of social biases while maintaining the V-L performance. 

\section{Conclusion}
This work presents a comprehensive study of social bias removal in CLIP. We first revisit several existing methods and point out their over-debiasing problem. We then leverage probing tests to illuminate the innate reason for this problem. In light of our preliminary findings, we propose a debiasing method that aligns the biases in embeddings, accompanied by delicately designed counterfactual debiasing. To fairly evaluate the bias removal effect, we also advocate two evaluation protocols that involve a new holistic metric and a generalization test operation. Our experiments demonstrate that the proposed method achieves a better trade-off between CLIP model debiasing and V-L alignment.

{
    \small
    \bibliographystyle{ieeenat_fullname}
    \bibliography{main}

\begin{thebibliography}{58}
\providecommand{\natexlab}[1]{#1}
\providecommand{\url}[1]{\texttt{#1}}
\expandafter\ifx\csname urlstyle\endcsname\relax
  \providecommand{\doi}[1]{doi: #1}\else
  \providecommand{\doi}{doi: \begingroup \urlstyle{rm}\Url}\fi

\bibitem[Agarwal et~al.(2021)Agarwal, Krueger, Clark, Radford, Kim, and Brundage]{agarwal-2021-evaluatingclip}
Sandhini Agarwal, Gretchen Krueger, Jack Clark, Alec Radford, Jong~Wook Kim, and Miles Brundage.
\newblock Evaluating clip: Towards characterization of broader capabilities and downstream implications, 2021.

\bibitem[Alvi et~al.(2019)Alvi, Zisserman, and Nell\r{a}ker]{alvi-2019-confusion_loss}
Mohsan Alvi, Andrew Zisserman, and Christoffer Nell\r{a}ker.
\newblock Turning a blind eye: Explicit removal of biases and variation from deep neural network embeddings.
\newblock In \emph{Computer Vision – ECCV 2018 Workshops}, page 556–572. Springer-Verlag, 2019.

\bibitem[Berg et~al.(2022)Berg, Hall, Bhalgat, Kirk, Shtedritski, and Bain]{berg-etal-2022-prompt}
Hugo Berg, Siobhan Hall, Yash Bhalgat, Hannah Kirk, Aleksandar Shtedritski, and Max Bain.
\newblock A prompt array keeps the bias away: Debiasing vision-language models with adversarial learning.
\newblock In \emph{ACL}, pages 806--822. ACL, 2022.

\bibitem[Birhane et~al.(2021)Birhane, Prabhu, and Kahembwe]{birhane-2021-multimodal}
Abeba Birhane, Vinay~Uday Prabhu, and Emmanuel Kahembwe.
\newblock Multimodal datasets: misogyny, pornography, and malignant stereotypes, 2021.

\bibitem[Bolukbasi et~al.(2016)Bolukbasi, Chang, Zou, Saligrama, and Kalai]{bolu-2016-word_projection}
Tolga Bolukbasi, Kai-Wei Chang, James Zou, Venkatesh Saligrama, and Adam Kalai.
\newblock Man is to computer programmer as woman is to homemaker? debiasing word embeddings.
\newblock In \emph{NIPS}, page 4356–4364. Curran Associates Inc., 2016.

\bibitem[Caliskan et~al.(2017)Caliskan, Bryson, and Narayanan]{caliskan-2017-weat}
Aylin Caliskan, Joanna~J. Bryson, and Arvind Narayanan.
\newblock Semantics derived automatically from language corpora contain human-like biases.
\newblock \emph{Science}, \penalty0 (6334):\penalty0 183--186, 2017.

\bibitem[Cheng et~al.(2021)Cheng, Hao, Yuan, Si, and Carin]{pengyu-2021-fairfil}
Pengyu Cheng, Weituo Hao, Siyang Yuan, Shijing Si, and Lawrence Carin.
\newblock Fairfil: Contrastive neural debiasing method for pretrained text encoders.
\newblock In \emph{ICLR}. OpenReview.net, 2021.

\bibitem[Chuang et~al.(2023)Chuang, Jampani, Li, Torralba, and Jegelka]{chuang-2023-debiasing}
Ching-Yao Chuang, Varun Jampani, Yuanzhen Li, Antonio Torralba, and Stefanie Jegelka.
\newblock Debiasing vision-language models via biased prompts, 2023.

\bibitem[Dehdashtian et~al.(2024{\natexlab{a}})Dehdashtian, Sadeghi, and Boddeti]{Dehdashtian_2024_CVPR}
Sepehr Dehdashtian, Bashir Sadeghi, and Vishnu~Naresh Boddeti.
\newblock Utility-fairness trade-offs and how to find them.
\newblock In \emph{CVPR}, pages 12037--12046, 2024{\natexlab{a}}.

\bibitem[Dehdashtian et~al.(2024{\natexlab{b}})Dehdashtian, Wang, and Boddeti]{dehdashtian-2024-fairerclip}
Sepehr Dehdashtian, Lan Wang, and Vishnu Boddeti.
\newblock Fairer{CLIP}: Debiasing {CLIP}'s zero-shot predictions using functions in {RKHS}s.
\newblock In \emph{ICLR}, 2024{\natexlab{b}}.

\bibitem[Deng et~al.(2009)Deng, Dong, Socher, Li, Li, and Fei-Fei]{deng-2009-imagenet}
Jia Deng, Wei Dong, Richard Socher, Li-Jia Li, Kai Li, and Li Fei-Fei.
\newblock Imagenet: A large-scale hierarchical image database.
\newblock In \emph{CVPR}, pages 248--255, 2009.

\bibitem[Devlin et~al.(2019)Devlin, Chang, Lee, and Toutanova]{devlin-etal-2019-bert}
Jacob Devlin, Ming-Wei Chang, Kenton Lee, and Kristina Toutanova.
\newblock {BERT}: Pre-training of deep bidirectional transformers for language understanding.
\newblock In \emph{NAACL}, pages 4171--4186. ACL, 2019.

\bibitem[Du et~al.(2021)Du, Mukherjee, Wang, Tang, Awadallah, and Hu]{mengnan-2021-fairness_module}
Mengnan Du, Subhabrata Mukherjee, Guanchu Wang, Ruixiang Tang, Ahmed Awadallah, and Xia Hu.
\newblock Fairness via representation neutralization.
\newblock In \emph{NIPS}, pages 12091--12103. Curran Associates, Inc., 2021.

\bibitem[Feldman et~al.(2015)Feldman, Friedler, Moeller, Scheidegger, and Venkatasubramanian]{feldman-2015-dp}
Michael Feldman, Sorelle~A. Friedler, John Moeller, Carlos Scheidegger, and Suresh Venkatasubramanian.
\newblock Certifying and removing disparate impact.
\newblock In \emph{KDD}, page 259–268. ACM, 2015.

\bibitem[Garcia et~al.(2023)Garcia, Hirota, Wu, and Nakashima]{Garcia_2023_CVPR}
Noa Garcia, Yusuke Hirota, Yankun Wu, and Yuta Nakashima.
\newblock Uncurated image-text datasets: Shedding light on demographic bias.
\newblock In \emph{Proceedings of the IEEE/CVF Conference on Computer Vision and Pattern Recognition (CVPR)}, pages 6957--6966, 2023.

\bibitem[Gerych et~al.(2024)Gerych, Zhang, Hamidieh, Pan, Sharma, Hartvigsen, and Ghassemi]{gerych-2024-bendvlm}
Walter Gerych, Haoran Zhang, Kimia Hamidieh, Eileen Pan, Maanas Sharma, Thomas Hartvigsen, and Marzyeh Ghassemi.
\newblock Bend{VLM}: Test-time debiasing of vision-language embeddings.
\newblock In \emph{NIPS}, 2024.

\bibitem[Geyik et~al.(2019)Geyik, Ambler, and Kenthapadi]{geyik-2019-linkedin}
Sahin~Cem Geyik, Stuart Ambler, and Krishnaram Kenthapadi.
\newblock Fairness-aware ranking in search \& recommendation systems with application to linkedin talent search.
\newblock In \emph{KDD}, page 2221–2231. ACM, 2019.

\bibitem[Gustafson et~al.(2023)Gustafson, Rolland, Ravi, Duval, Adcock, Fu, Hall, and Ross]{gusta-2023-facet}
Laura Gustafson, Chloe Rolland, Nikhila Ravi, Quentin Duval, Aaron Adcock, Cheng-Yang Fu, Melissa Hall, and Candace Ross.
\newblock Facet: Fairness in computer vision evaluation benchmark.
\newblock In \emph{ICCV}, pages 20313--20325, 2023.

\bibitem[Hardt et~al.(2016)Hardt, Price, and Srebro]{hardt-2016-eo}
Moritz Hardt, Eric Price, and Nathan Srebro.
\newblock Equality of opportunity in supervised learning.
\newblock In \emph{NIPS}, page 3323–3331. Curran Associates Inc., 2016.

\bibitem[He et~al.(2020)He, Fan, Wu, Xie, and Girshick]{He-2020-moco}
Kaiming He, Haoqi Fan, Yuxin Wu, Saining Xie, and Ross Girshick.
\newblock Momentum contrast for unsupervised visual representation learning.
\newblock In \emph{CVPR}, pages 9726--9735, 2020.

\bibitem[Hirota et~al.(2024)Hirota, Chen, Wang, Nakashima, Wang, and Hachiuma]{hirota-2024-saner}
Yusuke Hirota, Min-Hung Chen, Chien-Yi Wang, Yuta Nakashima, Yu-Chiang~Frank Wang, and Ryo Hachiuma.
\newblock Saner: Annotation-free societal attribute neutralizer for debiasing clip, 2024.

\bibitem[Kim et~al.(2019)Kim, Ghorbani, and Zou]{kim-2019-multiacc}
Michael~P. Kim, Amirata Ghorbani, and James Zou.
\newblock Multiaccuracy: Black-box post-processing for fairness in classification.
\newblock In \emph{Proceedings of the 2019 AAAI/ACM Conference on AI, Ethics, and Society}, page 247–254. ACM, 2019.

\bibitem[Kirichenko et~al.(2023)Kirichenko, Izmailov, and Wilson]{polina-2023-lastlayer}
Polina Kirichenko, Pavel Izmailov, and Andrew~Gordon Wilson.
\newblock Last layer re-training is sufficient for robustness to spurious correlations.
\newblock In \emph{ICLR}. OpenReview.net, 2023.

\bibitem[Kong et~al.(2024)Kong, Yuan, Hao, and Henao]{kong-2024-ttbias}
Fanjie Kong, Shuai Yuan, Weituo Hao, and Ricardo Henao.
\newblock Mitigating test-time bias for fair image retrieval.
\newblock In \emph{NIPs}. Curran Associates Inc., 2024.

\bibitem[Kärkkäinen and Joo(2021)]{kar-2021-fairface}
Kimmo Kärkkäinen and Jungseock Joo.
\newblock Fairface: Face attribute dataset for balanced race, gender, and age for bias measurement and mitigation.
\newblock In \emph{WACV}, pages 1547--1557, 2021.

\bibitem[Li et~al.(2021{\natexlab{a}})Li, Selvaraju, Gotmare, Joty, Xiong, and Hoi]{li-2021-albef}
Junnan Li, Ramprasaath Selvaraju, Akhilesh Gotmare, Shafiq Joty, Caiming Xiong, and Steven Chu~Hong Hoi.
\newblock Align before fuse: Vision and language representation learning with momentum distillation.
\newblock In \emph{NIPS}, pages 9694--9705. Curran Associates, Inc., 2021{\natexlab{a}}.

\bibitem[Li et~al.(2022)Li, Li, Xiong, and Hoi]{junnan-2022-blip}
Junnan Li, Dongxu Li, Caiming Xiong, and Steven C.~H. Hoi.
\newblock {BLIP:} bootstrapping language-image pre-training for unified vision-language understanding and generation.
\newblock In \emph{ICML}, pages 12888--12900. {PMLR}, 2022.

\bibitem[Li et~al.(2021{\natexlab{b}})Li, Gong, Liu, Wang, Qiao, and Cheng]{li-2021-longtailaug}
Shuang Li, Kaixiong Gong, Chi~Harold Liu, Yulin Wang, Feng Qiao, and Xinjing Cheng.
\newblock Metasaug: Meta semantic augmentation for long-tailed visual recognition.
\newblock In \emph{CVPR}, pages 5208--5217, 2021{\natexlab{b}}.

\bibitem[Liang et~al.(2020)Liang, Li, Zheng, Lim, Salakhutdinov, and Morency]{liang-etal-2020-sentDebias}
Paul~Pu Liang, Irene~Mengze Li, Emily Zheng, Yao~Chong Lim, Ruslan Salakhutdinov, and Louis-Philippe Morency.
\newblock Towards debiasing sentence representations.
\newblock In \emph{ACL}, pages 5502--5515. ACL, 2020.

\bibitem[Liu et~al.(2021)Liu, Haghgoo, Chen, Raghunathan, Koh, Sagawa, Liang, and Finn]{liu-2021-traintwice}
Evan~Z Liu, Behzad Haghgoo, Annie~S Chen, Aditi Raghunathan, Pang~Wei Koh, Shiori Sagawa, Percy Liang, and Chelsea Finn.
\newblock Just train twice: Improving group robustness without training group information.
\newblock In \emph{ICML}, pages 6781--6792. PMLR, 2021.

\bibitem[May et~al.(2019)May, Wang, Bordia, Bowman, and Rudinger]{may-etal-2019-seat}
Chandler May, Alex Wang, Shikha Bordia, Samuel~R. Bowman, and Rachel Rudinger.
\newblock On measuring social biases in sentence encoders.
\newblock In \emph{NAACL}, pages 622--628. ACL, 2019.

\bibitem[McCoy et~al.(2019)McCoy, Pavlick, and Linzen]{mccoy-etal-2019-right}
Tom McCoy, Ellie Pavlick, and Tal Linzen.
\newblock Right for the wrong reasons: Diagnosing syntactic heuristics in natural language inference.
\newblock In \emph{ACL}, pages 3428--3448. ACL, 2019.

\bibitem[Menon et~al.(2021)Menon, Jayasumana, Rawat, Jain, Veit, and Kumar]{aditya-2021-logitadj}
Aditya~Krishna Menon, Sadeep Jayasumana, Ankit~Singh Rawat, Himanshu Jain, Andreas Veit, and Sanjiv Kumar.
\newblock Long-tail learning via logit adjustment.
\newblock In \emph{ICLR}. OpenReview.net, 2021.

\bibitem[Nadeem et~al.(2021)Nadeem, Bethke, and Reddy]{nadeem-etal-2021-stereoset}
Moin Nadeem, Anna Bethke, and Siva Reddy.
\newblock {S}tereo{S}et: Measuring stereotypical bias in pretrained language models.
\newblock In \emph{ACL}, pages 5356--5371. ACL, 2021.

\bibitem[Nan et~al.(2021)Nan, Zeng, Qiao, Guo, and Lu]{nan-etal-2021-uncovering}
Guoshun Nan, Jiaqi Zeng, Rui Qiao, Zhijiang Guo, and Wei Lu.
\newblock Uncovering main causalities for long-tailed information extraction.
\newblock In \emph{EMNLP}, pages 9683--9695. ACL, 2021.

\bibitem[Nangia et~al.(2020)Nangia, Vania, Bhalerao, and Bowman]{nangia-etal-2020-crows}
Nikita Nangia, Clara Vania, Rasika Bhalerao, and Samuel~R. Bowman.
\newblock {C}row{S}-pairs: A challenge dataset for measuring social biases in masked language models.
\newblock In \emph{EMNLP}, pages 1953--1967. ACL, 2020.

\bibitem[Niven and Kao(2019)]{niven-kao-2019-probing}
Timothy Niven and Hung-Yu Kao.
\newblock Probing neural network comprehension of natural language arguments.
\newblock In \emph{ACL}, pages 4658--4664. ACL, 2019.

\bibitem[Plummer et~al.(2015)Plummer, Wang, Cervantes, Caicedo, Hockenmaier, and Lazebnik]{plummer-2015-flickr}
Bryan~A. Plummer, Liwei Wang, Chris~M. Cervantes, Juan~C. Caicedo, Julia Hockenmaier, and Svetlana Lazebnik.
\newblock Flickr30k entities: Collecting region-to-phrase correspondences for richer image-to-sentence models.
\newblock In \emph{ICCV}, pages 2641--2649, 2015.

\bibitem[Radford et~al.(2019)Radford, Wu, Child, Luan, Amodei, and Sutskever]{radford-2019-gpt}
Alec Radford, Jeffrey Wu, Rewon Child, David Luan, Dario Amodei, and Ilya Sutskever.
\newblock Language models are unsupervised multitask learners.
\newblock 2019.

\bibitem[Radford et~al.(2021)Radford, Kim, Hallacy, Ramesh, Goh, Agarwal, Sastry, Askell, Mishkin, Clark, Krueger, and Sutskever]{Radford-2021-clip}
Alec Radford, Jong~Wook Kim, Chris Hallacy, Aditya Ramesh, Gabriel Goh, Sandhini Agarwal, Girish Sastry, Amanda Askell, Pamela Mishkin, Jack Clark, Gretchen Krueger, and Ilya Sutskever.
\newblock Learning transferable visual models from natural language supervision.
\newblock In \emph{ICML}, pages 8748--8763. {PMLR}, 2021.

\bibitem[Ravfogel et~al.(2020)Ravfogel, Elazar, Gonen, Twiton, and Goldberg]{ravfogel-etal-2020-inlp}
Shauli Ravfogel, Yanai Elazar, Hila Gonen, Michael Twiton, and Yoav Goldberg.
\newblock Null it out: Guarding protected attributes by iterative nullspace projection.
\newblock In \emph{ACL}, pages 7237--7256. ACL, 2020.

\bibitem[Schick et~al.(2021)Schick, Udupa, and Schütze]{schick-2021-self-debias}
Timo Schick, Sahana Udupa, and Hinrich Schütze.
\newblock {Self-Diagnosis and Self-Debiasing: A Proposal for Reducing Corpus-Based Bias in NLP}.
\newblock \emph{ACL}, pages 1408--1424, 2021.

\bibitem[Seth et~al.(2023)Seth, Hemani, and Agarwal]{seth-2023-dear}
Ashish Seth, Mayur Hemani, and Chirag Agarwal.
\newblock Dear: Debiasing vision-language models with additive residuals.
\newblock In \emph{CVPR}, pages 6820--6829, 2023.

\bibitem[Singla et~al.(2022)Singla, Moayeri, and Feizi]{singla-2022-core}
Sahil Singla, Mazda Moayeri, and Soheil Feizi.
\newblock Core risk minimization using salient imagenet, 2022.

\bibitem[Steed and Caliskan(2021)]{steed-2021-ieat}
Ryan Steed and Aylin Caliskan.
\newblock Image representations learned with unsupervised pre-training contain human-like biases.
\newblock In \emph{FAccT}, page 701–713. ACM, 2021.

\bibitem[Tsirigotis et~al.(2023)Tsirigotis, Monteiro, Rodriguez, Vazquez, and Courville]{tsir-2023-withoutgroup}
Christos Tsirigotis, Joao Monteiro, Pau Rodriguez, David Vazquez, and Aaron~C Courville.
\newblock Group robust classification without any group information.
\newblock In \emph{NIPS}, pages 56553--56575. Curran Associates, Inc., 2023.

\bibitem[Wang et~al.(2021)Wang, Liu, and Wang]{wang-etal-2021-gender}
Jialu Wang, Yang Liu, and Xin Wang.
\newblock Are gender-neutral queries really gender-neutral? mitigating gender bias in image search.
\newblock In \emph{EMNLP}, pages 1995--2008. ACL, 2021.

\bibitem[Wang et~al.(2022{\natexlab{a}})Wang, Liu, and Wang]{wang-etal-2022-assessing}
Jialu Wang, Yang Liu, and Xin Wang.
\newblock Assessing multilingual fairness in pre-trained multimodal representations.
\newblock In \emph{ACL 2022}, pages 2681--2695. ACL, 2022{\natexlab{a}}.

\bibitem[Wang et~al.(2022{\natexlab{b}})Wang, Zhang, and Sang]{wang-2022-fairclip}
Junyang Wang, Yi Zhang, and Jitao Sang.
\newblock Fairclip: Social bias elimination based on attribute prototype learning and representation neutralization, 2022{\natexlab{b}}.

\bibitem[Wang et~al.(2019)Wang, Zhao, Yatskar, Chang, and Ordonez]{tianlu-2019-adversarial}
Tianlu Wang, Jieyu Zhao, Mark Yatskar, Kai-Wei Chang, and Vicente Ordonez.
\newblock Balanced datasets are not enough: Estimating and mitigating gender bias in deep image representations.
\newblock In \emph{ICCV}, pages 5309--5318, 2019.

\bibitem[Webster et~al.(2021)Webster, Wang, Tenney, Beutel, Pitler, Pavlick, Chen, Chi, and Petrov]{kellie-2021-dropout_debias}
Kellie Webster, Xuezhi Wang, Ian Tenney, Alex Beutel, Emily Pitler, Ellie Pavlick, Jilin Chen, Ed Chi, and Slav Petrov.
\newblock Measuring and reducing gendered correlations in pre-trained models, 2021.

\bibitem[Wolfe et~al.(2022)Wolfe, Banaji, and Caliskan]{wolfe-2022-evidence}
Robert Wolfe, Mahzarin~R. Banaji, and Aylin Caliskan.
\newblock Evidence for hypodescent in visual semantic ai.
\newblock In \emph{FaccT '22}, page 1293–1304. ACM, 2022.

\bibitem[Wortsman et~al.(2022)Wortsman, Ilharco, Kim, Li, Kornblith, Roelofs, Lopes, Hajishirzi, Farhadi, Namkoong, and Schmidt]{Wortsman_2022_CVPR}
Mitchell Wortsman, Gabriel Ilharco, Jong~Wook Kim, Mike Li, Simon Kornblith, Rebecca Roelofs, Raphael~Gontijo Lopes, Hannaneh Hajishirzi, Ali Farhadi, Hongseok Namkoong, and Ludwig Schmidt.
\newblock Robust fine-tuning of zero-shot models.
\newblock In \emph{CVPR}, pages 7959--7971, 2022.

\bibitem[Zhang and R\'{e}(2022)]{zhang-2022-adapter}
Michael Zhang and Christopher R\'{e}.
\newblock Contrastive adapters for foundation model group robustness.
\newblock In \emph{NIPS}, pages 21682--21697. Curran Associates, Inc., 2022.

\bibitem[Zhang et~al.(2022)Zhang, Hooi, Hong, and Feng]{zhang-2022-longtail}
Yifan Zhang, Bryan Hooi, Lanqing Hong, and Jiashi Feng.
\newblock Self-supervised aggregation of diverse experts for test-agnostic long-tailed recognition.
\newblock In \emph{NIPS}. Curran Associates Inc., 2022.

\bibitem[Zhang et~al.(2017)Zhang, Song, and Qi]{zhang-2017-utkface}
Z. Zhang, Y. Song, and H. Qi.
\newblock Age progression/regression by conditional adversarial autoencoder.
\newblock In \emph{CVPR}, pages 4352--4360. IEEE Computer Society, 2017.

\bibitem[Zhong et~al.(2022)Zhong, Yang, and Xu]{zhong-etal-2022-reducing}
Zeyi Zhong, Min Yang, and Ruifeng Xu.
\newblock Reducing spurious correlations for answer selection by feature decorrelation and language debiasing.
\newblock In \emph{COLING}, pages 1753--1764. International Committee on Computational Linguistics, 2022.

\bibitem[Zhu et~al.(2023)Zhu, Niu, Lee, Hur, and Zhang]{zhu-2023-finetune}
Beier Zhu, Yulei Niu, Saeil Lee, Minhoe Hur, and Hanwang Zhang.
\newblock Debiased fine-tuning for vision-language models by prompt regularization.
\newblock In \emph{AAAI}. AAAI Press, 2023.

\end{thebibliography}
}


\end{document}